\documentclass[twoside]{article}
\usepackage[accepted]{aistats2026}
\usepackage[round]{natbib}
\renewcommand{\bibname}{References}
\renewcommand{\bibsection}{\subsubsection*{\bibname}}

\usepackage{xcolor}
\usepackage{subcaption}
\usepackage{hyperref}
\hypersetup{
    colorlinks,
    linkcolor={black},
    citecolor={black},
    urlcolor={black}
}
\usepackage{soul}

\usepackage{amsthm}
\usepackage{cleveref}

\usepackage[utf8]{inputenc}
\usepackage[T1]{fontenc}

\usepackage{hyperref}    
\usepackage{xr-hyper}    

\usepackage{xcite}       
\usepackage{url}
\usepackage{booktabs}
\usepackage{array}
\usepackage{amsfonts}
\usepackage{nicefrac}
\usepackage{microtype}
\usepackage{xcolor}
\usepackage{amsmath}
\usepackage{enumitem}
\usepackage{multirow}
\usepackage{tablefootnote}
\usepackage{bm}
\usepackage{bbm}
\usepackage{amssymb}
\usepackage{xspace}
\usepackage{tikz}
\usepackage[colorinlistoftodos]{todonotes}
\usepackage{pgfplots}
\usepgfplotslibrary{groupplots}
\pgfplotsset{compat=1.17}
\usepackage{graphicx}
\usepackage{subcaption}
\usepackage{float}
\usepackage{wrapfig}
\usepackage{tcolorbox}
\usepackage[normalem]{ulem}
\usepackage{pdfpages}
\usepackage{soul}
\usepackage[round]{natbib}
\usepackage{cleveref}

\newtheorem{theorem}{\bf{Theorem}}
\newtheorem{lemma}{\bf{Lemma}}

\newtheorem{definition}{\bf{Definition}}

\newenvironment{theorem*}[1][]{%
    \par\addvspace{\topsep}
    \noindent{\bfseries Theorem\ifx&#1&\relax\else\ (#1)\fi\quad}\itshape
    \ignorespaces
}{%
    \par\addvspace{\topsep}
}

\newenvironment{lemma*}[1][]{%
    \par\addvspace{\topsep}
    \noindent{\bfseries Lemma\ifx&#1&\relax\else\ (#1)\fi\quad}\itshape
    \ignorespaces
}{%
    \par\addvspace{\topsep}
}

\newenvironment{proposition*}[1][]{%
    \par\addvspace{\topsep}
    \noindent{\bfseries Proposition\ifx&#1&\relax\else\ (#1)\fi\quad}\itshape
    \ignorespaces
}{%
    \par\addvspace{\topsep}
}

\newenvironment{corollary*}[1][]{%
    \par\addvspace{\topsep}
    \noindent{\bfseries Corollary\ifx&#1&\relax\else\ (#1)\fi\quad}\itshape
    \ignorespaces
    }{%
    \par\addvspace{\topsep}
}

\newenvironment{remark*}[1][]{%
    \par\addvspace{\topsep}
    \noindent{\bfseries Remark\ifx&#1&\relax\else\ (#1)\fi\quad}\itshape
    \ignorespaces
}{%
    \par\addvspace{\topsep}
}

\DeclareMathAlphabet{\altmathcal}{OMS}{cmsy}{m}{n} 

\newcommand{\tr}{\operatorname{tr}}

\newcommand{\xmath}[1] {\ensuremath{#1}\xspace}
\newcommand{\blmath}[1] {\xmath{\bm{#1}}}
\newcommand{\A}{\blmath{A}}

\newcommand{\Hmat}{\blmath{H}}
\newcommand{\I}{\blmath{I}}

\newcommand{\U}{\blmath{U}}
\newcommand{\V}{\blmath{V}}
\newcommand{\W}{\blmath{W}}
\newcommand{\X}{\blmath{X}}
\newcommand{\Y}{\blmath{Y}}
\newcommand{\Z}{\blmath{Z}}

\newcommand{\vzeta}{\blmath{\zeta}}
\newcommand{\vgamma}{\blmath{\gamma}}

\newcommand{\vginit}{\blmath{\gamma}^{\texttt{init}}}
\newcommand{\vgdebias}{\hat{\blmath{\gamma}}^{\mathtt{calib}}}

\newcommand{\hU}{\hat{\U}}

\renewcommand{\mathbf}{\boldsymbol}

\usepackage{xspace}

\newcommand{\norm}[1]{\left\lVert#1\right\rVert}

\newcommand{\calO}{\mathcal{O}}

\newcommand{\calR}{\mathcal{R}}

\newcommand{\calZ}{\mathcal{Z}}

\newcommand{\BC}{\mathbb{C}}

\newcommand{\BE}{\mathbb{E}}

\newcommand{\BP}{\mathbb{P}}

\newcommand{\BR}{\mathbb{R}}

\newcommand{\bQ}{\mathbf{Q}}

\newcommand{\veca}{\mathbf{a}}
\newcommand{\vecb}{\mathbf{b}}

\newcommand{\vecw}{\mathbf{w}}
\newcommand{\vecx}{\mathbf{x}}
\newcommand{\vecy}{\mathbf{y}}
\newcommand{\vecz}{\mathbf{z}}

\newcommand{\tred}{\textcolor{red}}

\newcommand{\bsigmaone}{\hat{\bm{\Sigma}}_1}
\newcommand{\bsigmatwo}{\hat{\bm{\Sigma}}_2}

\begin{document}

%

%

\twocolumn[

\aistatstitle{Calibrated Principal Component Regression}

\aistatsauthor{ Yixuan Florence Wu \And Yilun Zhu \And  Lei Cao \And  Naichen Shi}


\aistatsaddress{ Northwestern University \And  University of Michigan \And Northwestern University \And Northwestern University} 


]

\begin{abstract}
We propose a new method for statistical inference in generalized linear models. In the overparameterized regime, Principal Component Regression (PCR) reduces variance by projecting high-dimensional data to a low-dimensional principal subspace before fitting. However, PCR incurs truncation bias whenever the true regression vector has mass outside the retained principal components (PC). 

To mitigate the bias, we propose Calibrated Principal Component Regression (CPCR), which first learns a low-variance prior in the PC subspace and then calibrates the model in the original feature space via a centered Tikhonov step. CPCR leverages cross-fitting and controls the truncation bias by softening PCR's hard cutoff. Theoretically, we calculate the out-of-sample risk in the random matrix regime, which shows that CPCR outperforms standard PCR when the regression signal has non-negligible components in low-variance directions. Empirically, CPCR consistently improves prediction across multiple overparameterized problems. The results highlight CPCR's stability and flexibility in modern overparameterized settings. 

\end{abstract}

\section{INTRODUCTION}\label{sec:intro}

In this paper, we consider the problem of overparameterized regression, where the number of variables exceeds the number of samples. Overparameterization has become commonplace across scientific and engineering domains. Advances in data acquisition technologies routinely generate high-dimensional data with rich structure. For instance, functional MRI (fMRI) studies may record thousands of features for only a few hundred subjects~\citep{casey2018adolescent}, and single-cell RNA sequencing can measure expression levels of $10^3$--$10^5$ genes across a comparable number of cells~\citep{wu2020tools}. On the other hand, the development of foundation models, such as large language models (LLMs)~\citep{radford2021learning} and vision foundation models ~\citep{siméoni2025dinov3}, produces rich embeddings up to thousands of dimensions that capture complex semantics and enable downstream tasks such as few-shot classification, inference, and reasoning~\citep{yuan2021florence, yu2022coca}. Together, these trends have created a new data landscape where having far more features than samples is increasingly common.

Building statistical models in the overparameterized regime is not easy. From an information-theoretic standpoint, regression in such regimes is inherently ill-posed, where infinitely many solutions can interpolate the training data \citep{mcrae2022harmless, tsigler2023benign, medvedev2024overfitting}, raising concerns about identifiability and generalization. 

Classical statistics resolves this ambiguity by imposing prior assumptions or structural constraints. Two canonical examples are LASSO~\citep{tibshirani1996regression} and ridge regression~\citep{hoerl1970ridge}, which employ an $\ell_1$ and an $\ell_2$ penalty, respectively. Beyond explicit regularization, recent work has revealed that optimization dynamics can induce implicit regularization. For example, early stopping approximates ridge regression~\citep{ali2019continuoustimeviewearlystopping}, and gradient descent in overparameterized models induces an implicit bias toward minimum-norm or low-complexity solutions~\citep{Hastie_Montanari_Rosset_Tibshirani_2022}.

From a Bayesian perspective, however, these methods correspond to simple priors that assume isotropic feature distributions and overlook the heteroskedasticity of real data. An alternative line of work focuses on learning low-dimensional latent representations of the covariates to enable a range of downstream analyses~\citep{wu2020tools, maulik2021latent, recanatesi2021predictive, zhang2024learning, bonneville2024comprehensive}. A classical statistical method along these lines is principal component regression (PCR), which combines unsupervised dimension reduction with regression by projecting the data onto its leading principal components prior to fitting the model. Implicitly, PCR assumes that a small number of latent factors govern the variability of both covariates and responses, allowing it to identify the relevant low-dimensional structure prior to regression.

However, PCR suffers from a fundamental limitation that it decouples feature extraction from prediction. This decoupling is problematic when the regression signal does not align with the leading principal components of the covariates. In such cases, the projections of regression coefficients on directions with low variance in covariates are discarded, which introduces \textbf{truncation bias}. As a result, PCR may substantially degrade.

To mitigate truncation bias while still leveraging latent structure, we propose \emph{Calibrated Principal Component Regression} (CPCR). CPCR introduces a sample-splitting and cross-fitting strategy: one split is used to learn a low-variance but potentially biased prior, and the other split calibrates regression coefficients in the full feature space. This two-stage design effectively corrects for bias introduced by discarding residual directions. 

\begin{figure}[ht]
    \centering
    \includegraphics[width=1\linewidth]{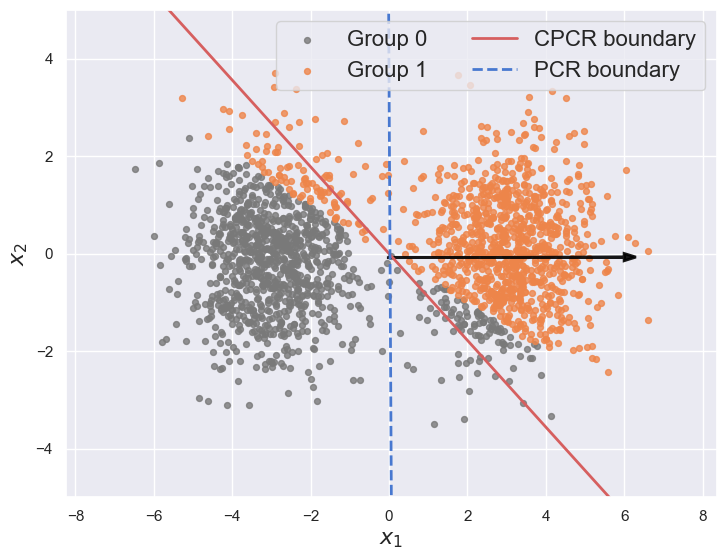}
    \caption{Data points with different membership, group 0 (grey dots) and group 1 (orange dots). The black arrow shows the top principal component. 
    When considering the top PC only, PCR's decision boundary is influenced by truncation bias, while CPCR is able to calibrate its decision boundary. }
    \label{fig:pcr_vs_cpcr}
\end{figure}

Figure~\ref{fig:pcr_vs_cpcr} illustrates this scenario in a 2D classification setting where the label signal is not perfectly aligned with the top principal component, and CPCR achieves a more accurate decision boundary than PCR.

Is the superior performance of CPCR merely an artifact of the specific dataset? Under what conditions does calibration provably mitigate truncation bias? To address these questions, we provide \textit{exact} theoretical characterizations of the out-of-sample prediction risk of CPCR under least-squares regression, leveraging tools from random matrix theory. 

We support these theoretical findings with extensive numerical experiments. The results consistently demonstrate that CPCR outperforms PCR, highlighting the benefit of calibration in reducing truncation bias. Moreover, CPCR is superior to ridge regression when the ground-truth regression vector is highly anisotropic and well-aligned with the principal subspace.  

On a wide range of overparameterized regression and classification problems, we evaluate CPCR with generalized linear models. Our results show consistent improvement over existing popular methods in high-dimensional regression.

In summary, our main contributions are:
\begin{enumerate}
    \item We present a novel algorithm, CPCR, that leverages cross-fitting to exploit the low-dimensional structure in covariate while controlling the truncation bias. 
    \item We develop new analytical tools to derive the \emph{exact} asymptotic risk of CPCR using random matrix theory. Our analysis quantifies how risk depends on (i) the alignment of the regression coefficient with the informative subspace and (ii) the spectral distribution of the covariance matrix constrained on the orthogonal subspace of principal components.
    \item We validate our theoretical risk characterization through synthetic experiments and demonstrate that CPCR consistently outperforms PCR and other popular baselines on both synthetic and real-world datasets.
\end{enumerate}

\subsection{Related Works}
We review several lines of related research most relevant to our framework of low-rank regression and calibration.
\paragraph{Principal component regression (PCR).}  PCR builds a regression model after projecting covariates onto leading principal components of the sample covariance. A recent line of work provides asymptotic and high-dimensional risk characterizations for PCR when the principal components are estimated from data rather than known a priori. In particular, \citet{2024nodoubledescent} provide asymptotic guarantees for generalization risk when data is sampled from a spiked covariance model, where the population covariance is nearly isotropic except for a few low-rank signal directions with elevated eigenvalues. Under a similar setting, \citet{green2024high} not only characterizes the exact asymptotic risk of PCR but also reveals the dependencies of optimal number of PCs. 
\citet{agarwal2023adaptive} propose an online variant of PCR with finite sample guarantees. Similar to ours, \citet{argawalrobustness} also considers the overparameterized regime. They study the robustness of PCR to noise, sparsity, and mixtures of the covariates. These works establish a solid theoretical foundations for PCR. Our contribution extends prior analyses by introducing a calibration step to standard PCR that mitigates its potential bias. 

\paragraph{Ridge regression in high dimensions.}
Ridge regression provides an isotropic prior to the regression coefficient by shrinking coefficients uniformly along all directions. \citet{bartlett2020benign} give necessary and sufficient spectral conditions under which ridgeless regression achieves nearly optimal risk in overparameterized linear models, thereby establishing when benign overfitting occurs. \citet{Hastie_Montanari_Rosset_Tibshirani_2022} characterize the generalization risk of ridge and ridgeless regression under isotropic and anisotropic features, and even for data transformed by a one-layer random neural network. \citet{bach2023highdimensionalanalysisdoubledescent} compute asymptotic equivalents for generalization performance when the data are first passed through a random projection matrix. More recently, \citet{cheng2025dimensionfreeridgeregression} establish non-asymptotic bias and variance bounds that hold beyond proportional asymptotics. Our theoretical analysis differs in two important ways. First, CPCR introduces a calibration step that solves a centered, regularized optimization problem, rather than relying on global isotropic shrinkage. Second, our theoretical analysis of asymptotic generalization risk extends beyond classical ridge regression by explicitly incorporating sample splitting and cross-fitting, which alters both the bias–variance tradeoff and the dependence on feature covariance structure.

\paragraph{Debiased regression and machine learning.} \citet{chernozhukov2018double} introduce a general framework that combines orthogonalization with cross-fitting to mitigate regularization bias when using machine learning in high-dimensional settings. To debias linear regression, \citet{yi2021new} propose a split-sample procedure, where the first half of the data is used to obtain a regularized estimator, and the second half is used to construct a debiased version without cross-fitting. Their abstract framework encompasses important classes of estimators, including the group LASSO and the Sorted L-One Penalized Estimator (SLOPE). Our debiasing scheme follows a similar split-then-debias principle, though specialized to principal component regression. In a broader perspective, debiasing has also been applied in other areas of machine learning, including semi-supervised learning, Gaussian processes, and prediction-powered inference \citep{schmutz2023dontfearunlabelledsafe, hu2022nonrandommissinglabelssemisupervised, ji2025predictionssurrogatesrevisitingsurrogate}. CPCR can be viewed as a specialized instance of debiased or calibrated estimation. Crucially, the structure of our problem allows us to derive stronger theoretical guarantees compared to general debiased machine learning frameworks.

\paragraph{Supervised dimension reduction.}
Partial least squares (PLS) and canonical-correlation analysis (CCA) are classical supervised linear methods that maximize covariance or correlation between projections of covariates and response. While CCA relies on inverting sample covariance matrices, which makes it unstable in high-dimensional regimes, PLS avoids this via iterative algorithms (e.g., NIPALS (Wold, 1975)). However, since both methods intrinsically rely on a small number of components, they are also subject to truncation bias. Sliced Inverse Regression (SIR)~\citep{li1991sliced} and Supervised PCA~\citep{bair2006prediction, ritchie2020supervised} construct supervised low-dimensional embeddings of covariates using nonlinear or semi-parametric transformations. More specifically, SIR identifies directions that explain how the inverse regression changes across slices of response, which estimates a low-dimensional subspace of the predictors relevant for predicting response. However, the standard form of SIR is known to be brittle in the overparameterized regime~\citep{li2008sliced}. On the other hand, supervised PCA finds principal components by weighting each feature according to its covariance with response. In contrast to these supervised approaches, CPCR builds on the unsupervised framework of PCR, introducing a calibration step within the regression stage rather than constructing response-dependent embeddings of the predictors.


\section{PROBLEM FORMULATION}\label{sec:formulation}



Our goal is to learn a predictive model that maps high-dimensional covariates to responses. Let $\X \in \mathbb{R}^{p \times n}$ denote the data matrix, where $p$ is the number of features and $n$ is the number of samples, and let $\vecy$ denote the $n$-dimensional response vector whose entries are continuous for regression tasks and categorical for classification tasks. We focus on the high-dimensional regime where $p$ is comparable to or larger than $n$. 

We model the conditional distribution of the response $\vecy$ given the covariates $\X$ using a generalized linear model (GLM):  
\begin{equation}\label{eq:cond}
    p(\vecy \mid \X) = \rho\!\left(\vecy;\, \X^\top \vgamma\right),
\end{equation}
where $\vgamma \in \mathbb{R}^p$ is the regression coefficient vector, and $\rho$ denotes a density or mass function from the exponential family parameterized by the linear predictor $\X^\top \vgamma$. Standard choices of $\rho$ include Gaussian and Bernoulli distributions, corresponding to least-squares linear and logistic regression, respectively.

We further consider the intrinsic structure of $\X$. A natural assumption is that the predictive signal in $\X$ can be well-approximated by a rank-$r$ subspace $\U \in \mathbb{R}^{p \times r}$, with $r \ll p$. Exploiting such low-rank structure is especially valuable in the overparameterized regime, where direct estimation in the ambient space suffers from high variance and instability. 

To quantify the alignment between the regression coefficient vector $\vgamma$ and the informative subspace, we introduce the following notion of a predictive-power coefficient.

\begin{definition}
Given a coefficient vector $\vgamma$ and a low-rank subspace $\U$, the \emph{predictive-power coefficient} $\kappa$ is defined as  
\begin{equation}\label{eq:kappa}
    \kappa \;=\; \frac{\norm{\bm{\Pi}_{\U}\vgamma}_2^2}{\norm{\vgamma}_2^2} \;\in\; (0,1),
\end{equation}
where $\bm{\Pi}_{\U}$ is the projection operator onto column space of $\U$, and $\|\cdot\|_2$ denotes the Euclidean norm.
\end{definition}

Intuitively, $\kappa$ measures the fraction of the regression signal explained by the subspace $\U$. Values close to $1$ indicate that the predictive signal is well captured by $\U$, while smaller values reflect stronger contributions from residual directions outside the subspace $\U$.  

When $\kappa$ is close to $1$, a natural strategy is principal component regression (PCR). Suppose we have an estimate $\hU \in \mathbb{R}^{p \times r}$ that approximates the column space of $\U$. We project the data onto this subspace to obtain low-dimensional embeddings $\hU^\top \X$, and solve the regression problem through $
    \min_{\vzeta} \; \mathcal{L}\bigl(\vecy,\, (\hU^\top \X)^\top \vzeta \bigr)$, 
where $\mathcal{L}$ denotes the negative log-likelihood loss function associated with the GLM in \eqref{eq:cond}. By projecting onto $\hU$, PCR reduces the effective degrees of freedom of the covariates, thereby reducing the variance of the estimates.




\section{METHOD}\label{sec:method}

In practice, the predictive-power coefficient $\kappa$ is rarely equal to $1$. Standard PCR therefore suffers from truncation bias, since the complementary component $(\I_p - \bm{\Pi}_{\U})\vgamma$ is completely discarded. When $\kappa$ is not close to $1$, ignoring these low-variance but informative directions can lead to substantial estimation error.  

To address this challenge, we propose a \emph{calibration-based} approach that combines dimension reduction with a debiasing step via sample splitting and cross-fitting. The procedure ensures that information in the discarded subspace is partially recovered, thereby mitigating truncation bias while retaining the variance reduction benefits of PCR.  

More specifically, we first randomly split the labeled dataset $(\X,\vecy)$ into two parts with equal size: $\X=\X_{1}\cup\X_{2}$ and $\vecy=\vecy_{1}\cup\vecy_{2}$. Next, we fit an estimator separately on each split and then combine them through calibration steps. We provide an explicit algorithm for computation described as below.

\paragraph{Step 1: PCR.} On $(\X_1,\vecy_1)$, we fit a PCR model and obtain:
\begin{equation}    
\hat{\vzeta}_1 = \arg\min_{\vzeta}\mathcal{L}(\vecy_1,(\hat{\U}^{\top}\X_1)^\top \vzeta)\, .
\end{equation}

\paragraph{Step 2: calibration.} We construct $\vgamma_1^{\mathtt{init}}$ as $\vgamma_1^{\mathtt{init}}=\hat{\U}\hat{\vzeta}_1$, with $\hat{\vzeta}_1$ estimated from the first step. Then, on the held-out split $(\X_2,\vecy_2)$, we calibrate $\vgamma$ as
    \begin{equation}    \label{eqn:debiasgeneral}\vgdebias_1{} = \arg\min_{\vgamma }\mathcal{L}(\vecy_2,\X_2^\top\vgamma)+\lambda\norm{\vgamma-\vgamma_1^{\mathtt{init}}}^2
    \end{equation}

The calibration step prevents radically misspecified $\vginit_1$ by regularizing it with the likelihood from the held-out set. 

\paragraph{Step 3: exchange and re-calibration.} Since we potentially only leverage the information from half of the samples in step 1 and 2, we exchange $\left(\X_1,\vecy_1\right)$ with $\left(\X_2,\vecy_2\right)$, and repeat step 1 and step 2 to obtain $\vgdebias_2{}$. Finally, our estimator for CPCR is:
\begin{equation}
    \label{eqn:debiasfinal}
    \hat{\vgamma}^{\mathtt{CPCR}} = \frac{1}{2}\left(\vgdebias_1{} + \vgdebias_2{}\right).
\end{equation}

The proposed CPCR estimator inherits the variance reduction benefits of PCR while correcting its truncation bias through calibration. The use of sample splitting and cross-fitting is essential in the overparameterized regime: without data splitting, PCR may interpolate the training set, leaving no residual signal for the calibration step to adjust, thereby rendering the procedure ineffective.


\section{ANALYSIS OF CALIBRATION ESTIMATOR RISK}
In this section, we develop the theoretical analysis of CPCR, providing an exact characterization of its generalization risk and identifying conditions under which it excels.

Our theoretical analysis builds on recent advances in random matrix theory for high-dimensional regression \citep{Dobriban_Wager_2018, Hastie_Montanari_Rosset_Tibshirani_2022, bach2023highdimensionalanalysisdoubledescent}, which provide asymptotic characterizations of generalization risk for classical estimators such as ridge regression and PCR. Distinct from these settings, CPCR introduces a sample-splitting calibration step, leading to nontrivial dependencies between intermediate estimators $\hat{\bm{\gamma}}^{\mathtt{calib}}_1$ and $\hat{\bm{\gamma}}^{\mathtt{calib}}_2$. This structure gives rise to additional cross terms that are absent in standard analyses. We develop new analytical tools to characterize or estimate the cross terms by combining deterministic functional calculus, spectral asymptotics, and high-dimensional concentration arguments.

\subsection{Risk Analysis of CPCR}
For clarity, we dedicate our analysis of CPCR to the setting of least-squares regression, a representative case that enables analytical tractability, while many key insights about calibration and truncation bias are transferable to broader GLMs. In this case, the GLM in \eqref{eq:cond} specializes to the Gaussian linear model. Equivalently,
\begin{equation}\label{eq:linear_model}
    \vecy=\X^\top\bm{\gamma}^*+\bm{\epsilon}\, ,
\end{equation}
where $\bm{\epsilon}\in \BR^{n\times1}$ is a noise vector with independent entries with mean zero and variance $\sigma^2$. The analysis in this section is based on model~\eqref{eq:linear_model}.

\paragraph{Data generation process.} We assume our covariate feature matrix $\X$ is generated by $\X = \bm{\Sigma}^{1/2}\Z \,$,
with $\Z \in \BR^{p \times n}$ having independent entries with zero mean and unit variance, and $\bm{\Sigma}\in\mathbb{R}^{p\times p}$ is a positive semidefinite population covariance matrix. 

We further consider a spiked model, where the population covariance matrix $\bm{\Sigma} \in \BR^{p \times p}$ can be written as 
\begin{equation*}
    \bm{\Sigma} = \U\bm{\Sigma}_s\U^\top + \V\bm{\Sigma}_c\V^\top \, ,
\end{equation*}
with entries of $\bm{\Sigma}_s \in \BR^{r\times r}$ being low rank signals and $\bm{\Sigma}_c \in \BR^{(p-r)\times (p-r)}$ represents the complementary background structure that is orthogonal to the signals space. Without loss of generality, both $\bm{\Sigma}_s$ and $\bm{\Sigma}_c$ are assumed to be diagonal. In addition, $\U \in \BR^{p\times r}$ and $\V \in \BR^{p\times (p-r)}$ are orthonormal factors with $\U^\top\U=\I_r$, $\V^\top\V=\I_{p-r}$ and $\U^\top \V=\mathbf{0}$.

\paragraph{Distributional assumptions on $\vgamma^*$.} 

In many applications, predictive information is concentrated in a small number of dominant directions, while the remaining directions contribute little signal. To capture this structure, we develop a prior distribution of $\vgamma^*$ that is compatible with the definition of $\kappa$ in~\eqref{eq:kappa}. 
\begin{align}\label{eq:gamma_cov}    \bm{\Sigma}_{\bm{\gamma}^*}=\kappa\bm{\Pi}_{\U}+(1-\kappa)\bm{\Pi}_{\V} \, .
\end{align}
One can verify that this modeling choice provides a versatile data generation process that aligns with the predictive-power coefficient $\kappa$.

\paragraph{Risk.} Given an estimator $\hat{\vgamma}$ fitted from $(\X,\vecy)$, we evaluate its performance on a new independent covariate $\vecx_0 = \bm{\Sigma}^{1/2}\vecz_0$ drawn from the population distribution. The out-of-sample prediction risk is defined as   
\begin{align}\label{eq:risk}
    \mathcal{R}(\hat{\vgamma})\overset{\Delta}{=}\BE_{\vecx_0,\vgamma^*}[ (\vecx_0^\top \hat{\vgamma} -  \vecx_0^\top \vgamma^*)^2 ] \, .
\end{align}

Though $\hat{\vgamma}^{\mathtt{CPCR}}$ is dependent on training set $(\X,\vecy)$, whose risk is a random variable, the following theorem presents a deterministic limit of $\mathcal{R}(\hat{\vgamma}^{\mathtt{CPCR}})$ in the large $n$ and $p$ regime.

\begin{theorem}[Exact calibration estimator risk.]\label{thrm:1}
In the asymptotic limit where $n,p \rightarrow\infty$ such that $p/n \rightarrow c\in(1,\infty)$. With high probability, the risk of our calibration estimator $\hat{\bm{\gamma}}^{\mathtt{calib}}$ converge to a deterministic limit: 
\begin{align}\label{eq:limit_theory}
    \mathcal{R}(\hat{\vgamma}^{\mathtt{CPCR}}) \rightarrow
    B_{\X}(\hat{\vgamma}^{\mathtt{CPCR}}) + V(\hat{\vgamma}^{\mathtt{CPCR}}) \, ,
\end{align}
where the bias term can be further estimated by
\begin{align*}
    B_{\X}(\hat{\vgamma}^{\mathtt{CPCR}}) = &\frac{1}{4}(B_{\X}(\hat{\bm{\gamma}}^{\mathtt{calib}}_1) + B_{\X}(\hat{\bm{\gamma}}^{\mathtt{calib}}_2)) \\+&\frac{1}{2}B_{\X}(\hat{\bm{\gamma}}^{\mathtt{calib}}_1, \hat{\bm{\gamma}}^{\mathtt{calib}}_2)
\end{align*}
and the variance term can be estimated by
\begin{align*}
     V(\hat{\vgamma}^{\mathtt{CPCR}}) = \frac{1}{4}(V_{\epsilon_2}(\hat{\bm{\gamma}}^{\mathtt{calib}}_1) + V_{\epsilon_1}(\hat{\bm{\gamma}}^{\mathtt{calib}}_2)) \, .
\end{align*}
Moreover, consider $\{\mu_j(\bm{\Sigma}_c)\}_{j=1}^{p-r}$ the diagonal entries of $\bm{\Sigma}_c$ and $\nu_c = \frac{1}{p-r}\sum_{j=1}^{p-r}\delta_{\mu_j(\bm{\Sigma}_c)}$, then the almost sure limits of each individual terms are
\begin{align}\label{eq:bias_1_limit}
    &B_{\X}(\hat{\bm{\gamma}}^{\mathtt{calib}}_1) \xrightarrow{\text{a.s.}}\\ 
    +&(1-\kappa)(p-r)\int\frac{\mu}{1+\Tilde{m}(-\lambda)\mu} d\nu_c(\mu) \nonumber\\
    +&(1-\kappa)(p-r)\Tilde{m}_\rho(-\lambda) \int\frac{\mu}{(1+\Tilde{m}(-\lambda)\mu)^2} d\nu_c(\mu)\nonumber\\
    -& (1-\kappa)(p-r)\Tilde{m}(-\lambda)\int\frac{\mu^2}{(1+\Tilde{m}(-\lambda)\mu)^2} d\nu_c(\mu) \nonumber  \, ,
\end{align}
\begin{align}\label{eq:bias_1_2_limit}
    &B_{\X}(\hat{\bm{\gamma}}^{\mathtt{calib}}_1, \hat{\bm{\gamma}}^{\mathtt{calib}}_2) \xrightarrow{\text{a.s.}} \\
    & (1-\kappa) (p-r)\int\frac{\mu}{(1+\Tilde{m}(-\lambda)\mu)^2} d\nu_c(\mu) \nonumber
\end{align} and
\begin{align} \label{eq:var_e2_limit}
    V_{\epsilon_2}(\hat{\bm{\gamma}}^{\mathtt{calib}}_1)  \xrightarrow{\text{a.s.}} \sigma^2  \left( \frac{\Tilde{m}_z(-\lambda)}{\Tilde{m}^2(-\lambda)}-1 \right). 
\end{align}
$\Tilde{m}(\cdot)$ is defined implicitly as the solutions to a nonlinear equation depending on $n, p$ and $\bm{\Sigma}$. $\Tilde{m}_\rho(\cdot)$ and $\Tilde{m}_z(\cdot)$ are partial derivatives with respect to $\rho$ and $z$ respectively. 
\end{theorem} 

From~\eqref{eq:limit_theory}, two main observations arise: (i) the risk decreases as $\kappa \to 1$, and (ii) the limiting risk depends only on $\nu_c$, the spectral distribution of the nuisance covariance $\bm{\Sigma}_c$, and not on the choice of basis $\U$ or $\V$. In addition, based on~\eqref{eq:limit_theory}, the following theorem specifies a condition for the optimal weight $\lambda$ in~\eqref{eqn:debiasgeneral}.



\begin{theorem}[Optimal $\lambda$.] \label{thrm:2}
    Under the same conditions as Theorem~\ref{thrm:1}, the optimal lambda $\lambda^*$ to achieve the minimal risk in (\ref{eq:risk}) satisfies the following equation:
    \begin{align*}
        B_{\X}'(\lambda^*)+V'(\lambda^*) = 0 \, .
    \end{align*}
    In particular, 
    \begin{align*}
        B_{\X}'(\lambda) =& \frac{(1-\kappa)(p-r)}{2} \biggl( 
        2\int\frac{\mu^2\Tilde{m}_z(-\lambda)}{(1+\Tilde{m}(-\lambda)\mu)^2} d\nu_c(\mu) \\     
        &- \int\frac{\mu\Tilde{m}_{\rho z}(-\lambda) }{(1+\Tilde{m}(-\lambda)\mu)^2} d\nu_c(\mu) \\       
        &+ 2 \int\frac{\mu^2(\Tilde{m}_{\rho}(-\lambda)\Tilde{m}_z(-\lambda)+\Tilde{m}_z(-\lambda))}{(1+\Tilde{m}(-\lambda)\mu)^3} d\nu_c(\mu) \\       
        &-  2\int\frac{\mu^3\Tilde{m}(-\lambda)\Tilde{m}_z(-\lambda)}{(1+\Tilde{m}(-\lambda)\mu)^3} d\nu_c(\mu) 
        \biggl)
    \end{align*}
    
    and
    \begin{align*}
         V'(\lambda) = \frac{\sigma^2}{2} \biggl( \frac{-\Tilde{m}(-\lambda)\Tilde{m}_{zz}(-\lambda)+2\Tilde{m}_z^2(-\lambda)}{\Tilde{m}^3(-\lambda)} \biggl) \, .
    \end{align*}
    $\Tilde{m}_{\rho z}(\cdot)$ denotes $\partial_{\rho z}\Tilde{m}(\cdot)$ and $\Tilde{m}_{z z}(\cdot)$ denotes $\partial_{zz}\Tilde{m}(\cdot)$. 
\end{theorem}
The proofs of Theorems~\ref{thrm:1} and~\ref{thrm:2}, along with the explicit forms of $\Tilde{m}(\cdot)$ and other auxiliary functions, are deferred to the Supplementary Materials. We will discuss the theoretical implications in the following sections.


\subsection{Numerical Validations}
\label{sec:verification}
\begin{figure}[ht]
    \centering
    \includegraphics[width=1\linewidth]{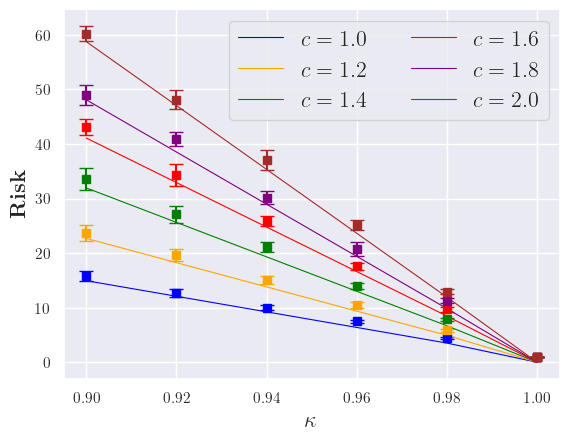}
    \caption{Theoretical limits of the empirical risks. Error bars denote the empirical risks (\ref{eq:risk}) and solid curves denote the theoretical limits (\ref{eq:limit_theory}). }
    \label{fig:theorem1}
\end{figure}

Theorem~\ref{thrm:1} provides several key theoretical insights into the risk behavior of CPCR. Before discussing these in detail, we first conduct numerical experiments to verify the asymptotic predictions. For each choice of aspect ratio $c$ and alignment parameter $\kappa$, we generate synthetic datasets following the process in Section~\ref{sec:method}, with $\vgamma^*$ sampled according to~\eqref{eq:gamma_cov}. Responses $\vecy$ are then generated using the linear model~\eqref{eq:linear_model}, yielding datasets $(\X,\vecy)$. We fit CPCR on each dataset to obtain $\hat{\vgamma}^{\mathtt{CPCR}}$ and compute empirical risks using~\eqref{eq:risk}. We leverage the true signal subspace for both CPCR and PCR in this section. Each experiment is repeated 10 times to construct error bars. For comparison, we calculate the theoretical limits predicted by Theorem~\ref{thrm:1}, where the optimal regularization parameter $\lambda^*$ is given by Theorem~\ref{thrm:2}. Figure~\ref{fig:theorem1} plots the empirical risks with error bars against the deterministic theoretical limits.  

Figure~\ref{fig:theorem1} shows close agreement between empirical risks and their theoretical counterparts across a range of aspect ratios $c$. The tight concentration of the error bars around the curves confirms that our asymptotic analysis accurately characterizes even finite-sample behavior. Moreover, Figure~\ref{fig:theorem1} illustrates the monotone decrease in risk as $\kappa$ increases. In particular, as $\kappa$ approaches $1$, the bias terms~\eqref{eq:bias_1_limit} and~\eqref{eq:bias_1_2_limit} completely vanish, leaving us with only the variance~\eqref{eq:var_e2_limit}.

\subsection{Implications of Theoretical Analysis}
 
\begin{figure}[ht]
    \centering
    \includegraphics[width=1\linewidth]{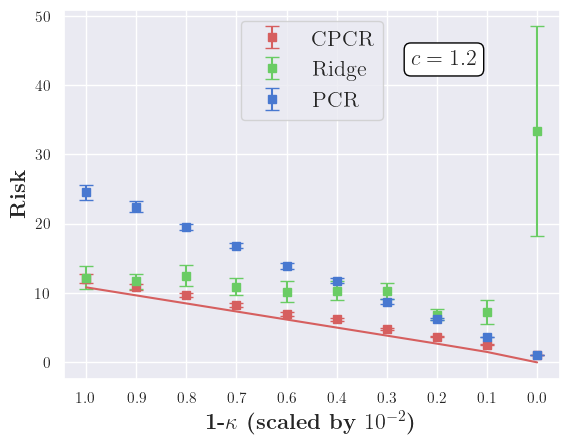}
    \caption{Risks of CPCR, PCR, and ridge regression. Eigenvalues of $\bm{\Sigma}_s$ are from $\left[2,4\right]$ and those of $\bm{\Sigma}_c$ are from $\left[1,3\right]$. CPCR consistently outperforms PCR and ridge regression. }
    \label{fig:cpcr_pcr_ridge_larger_V}
\end{figure}

\paragraph{Risk dependence on spectral distributions.} Since the bias term depends only on the eigenvalues of $\bm{\Sigma}_c$, we investigate how their magnitudes affect the risks. Specifically, we sample the entries of $\bm{\Sigma}_s$ from $\text{Unif}(2,4)$ and those of $\bm{\Sigma}_c$ from $\text{Unif}(1,3)$, where some eigenvalues of $\bm{\Sigma}_c$ are comparable to those of $\bm{\Sigma}_s$. In this regime, the signal and nuisance subspaces are not clearly separable (Figure~\ref{fig:cpcr_pcr_ridge_larger_V}). Several observations follow. First, CPCR consistently outperforms PCR except when $\kappa$ is nearly $1$, in which case they become comparable, which highlights the benefits of calibration. Second, both CPCR and PCR achieve low risk when $\kappa$ is close to $1$, indicating that they effectively leverage low-rank structure when the signals are concentrated in $\U$. Interestingly, ridge regression performs worse in this regime, with substantially higher variance. This occurs probably because ridge enforces an isotropic prior that shrinks all directions uniformly, penalizing even the signal-aligned components of $\vgamma^*$ and thereby inflating both bias and variance. Third, as $\kappa$ decreases, the risk of PCR grows rapidly due to truncation bias, whereas CPCR degrades more gracefully, which again confirms its robustness against truncation bias.


To further examine the sensitivity of the risks to the eigenvalues of $\bm{\Sigma}_c$, we vary the uniform sampling range of its entries so that the mean spans values between 1 and 6, while fixing the variance at $\tfrac{1}{3}$. The entries of $\bm{\Sigma}_s$ are independently drawn from $\text{Unif}(2,4)$, with mean 3 and variance $\tfrac{1}{3}$. 
 \begin{figure}[ht]
     \centering
     \includegraphics[width=1\linewidth]{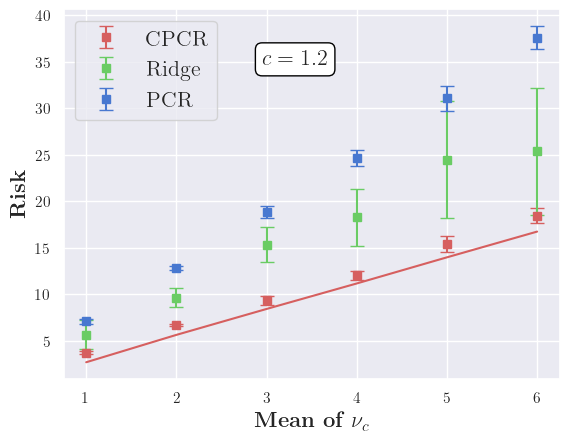}
     \caption{Risks of CPCR, PCR, and ridge regression with different range of eigenvalues in $\bm{\Sigma}_c$ with $\kappa=0.995$. CPCR is more resistant to the increasing magnitudes of signals in the background space.}
     \label{fig:diff_mean_sigma_c}
 \end{figure}
As is evident from Figure~\ref{fig:diff_mean_sigma_c}, CPCR is substantially more robust to increasing background variance. Its risk increases at a slower rate than that of ridge regression or PCR as the background signal strength increases.

\begin{figure}[ht]
    \centering
    \includegraphics[width=1\linewidth]{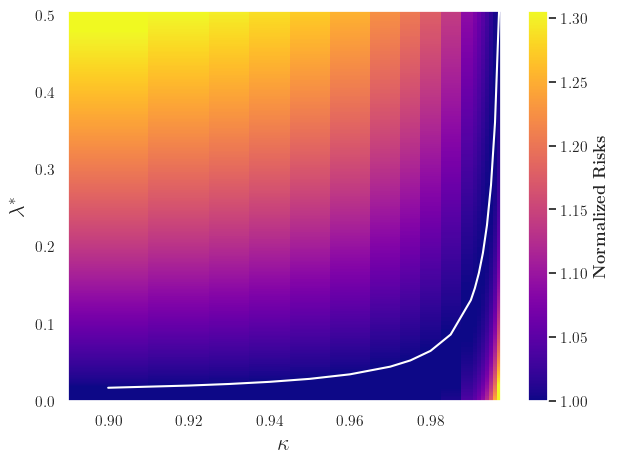}
    \caption{Normalized risk landscape over $(\kappa, \lambda)$. The white curve traces the optimal regularization parameter $\lambda^*$ for each $\kappa$. The background color represents the risk evaluated at each $(\kappa, \lambda)$ pair, divided by the optimal risk corresponding to that $\kappa$. Darker (purple) regions indicate risks closer to the optimum, while brighter (yellow) regions correspond to higher risks. }
    \label{fig:thrm2}
\end{figure}

\paragraph{Optimal $\lambda$.} As theorem~\ref{thrm:2} explains the relationship between the optimal calibration parameter $\lambda^*$ and the predictive-power coefficient $\kappa$ in a complicated manner, we plot $\lambda^*$ under different settings of $\kappa$ as the white curve in Figure~\ref{fig:thrm2}.

In Figure~\ref{fig:thrm2}, as $\kappa$ approaches $1$, $\lambda^*$ increases dramatically, indicating that the initial PCR estimate is already well aligned with the true regression coefficient, and additional calibration offers only marginal improvements. Conversely, for smaller values of $\kappa$, the optimal $\lambda^*$ becomes closer to zero. In such scenarios, the PCR solution does not offer strong guidance for estimating true regression coefficients, and the calibration step with a weaker regularization could significantly mitigate the truncation bias in the PCR solution. The behavior of optimal $\lambda^*$ is consistent with our intuitions.

\section{EXPERIMENTS}
We present experimental results on synthetic and real datasets, showing CPCR’s competitive or improved performance over benchmark methods across diverse tasks.

\subsection{Synthetic Data}

\begin{figure}[ht]
    \centering
    \includegraphics[width=1\linewidth]{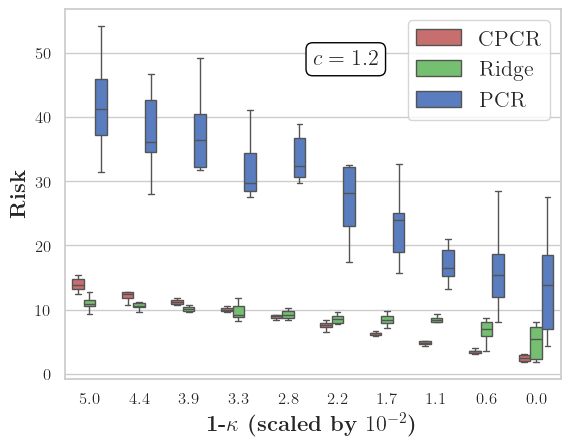}
    \caption{Risks of CPCR, ridge regression, and PCR when $\operatorname{rank}(\U)=10$ and $\operatorname{rank}(\hat{\U})=5$. CPCR consistently outperforms PCR. }
    \label{fig:cpcr_pcr_ridge}
\end{figure}

\paragraph{Data generation and methodology.} 
We compare the performance of CPCR with ridge regression and PCR on synthetic data. Following the procedure in Section~\ref{sec:method}, we generate a feature matrix $\X$ where the eigenvalues of $\bm{\Sigma}_c \sim \text{Unif}(0,1)$ are strictly smaller than those of $\bm{\Sigma}_s \sim \text{Unif}(2,4)$, ensuring a clear separation between signal and noise directions. The regression coefficients $\vgamma^*$ are sampled according to (\ref{eq:gamma_cov}), and responses $\vecy$ are generated from the linear model (\ref{eq:linear_model}) with additive Gaussian noise. To introduce estimator mis-specification, we perform a singular value decomposition of $\X$ and deliberately retain fewer principal components than the true signal rank (Figure~\ref{fig:cpcr_pcr_ridge}). PCR and CPCR are then applied in this reduced latent space, while ridge regression is fit in the original feature space. Empirical risks are evaluated via (\ref{eq:risk}), and the experiment is repeated 10 times, with results summarized in box plots.


\paragraph{Results.} 

It is evident that PCR suffers more severely from truncation bias when $\hat{\U}$ underestimates the rank, while CPCR is capable of calibrating on the misspecified subspace and successfully controlling that bias (Figure~\ref{fig:cpcr_pcr_ridge}). In addition, CPCR also achieves superior or comparable performance with ridge regression when $\kappa$ approaches 1. Our results demonstrate the robustness of CPCR in the presence of model misspecification.

\begin{figure}[ht]
    \centering
    \includegraphics[width=1\linewidth]{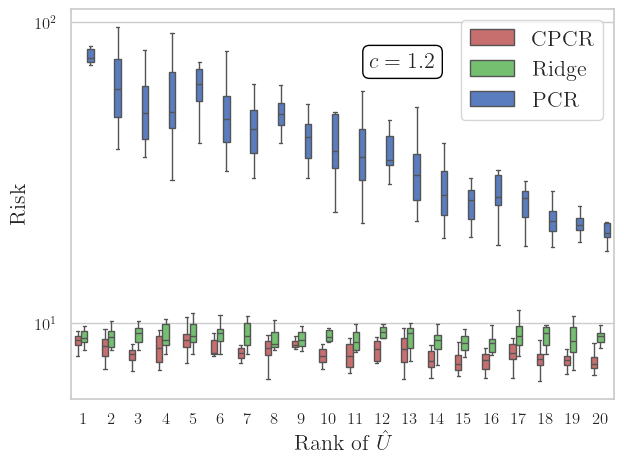}
    \caption{Empirical risks of CPCR, ridge regression and PCR against different $\operatorname{rank}(\hat{\U})$ with $c=1.2$ and $\kappa=0.98$. PCR benefits from more PC while CPCR and ridge regression is not sensitive to this parameter. }
    \label{fig:diff_r_comparing}
\end{figure}

To further investigate the estimator mis-specification and truncation bias, we vary $rank(\hat{\U})$ from $1$ to $20$ while keeping $rank(\U)=20$ and plot the risk of the three algorithms in Figure~\ref{fig:diff_r_comparing}.

Results from Figure~\ref{fig:diff_r_comparing} show that $\operatorname{rank}(\hat{\U})$ does not significantly affect the empirical risk of CPCR if $\kappa$ is large enough, again confirming the robustness of CPCR against misspscification. 

\subsection{Kernel Regression with Nystrom Features}
\label{subsec:exp_nystrom}

\paragraph{Datasets and Baselines.}
We evaluate our method on $5$ benchmark datasets from the UCI Machine Learning Repository~\citep{asuncion2007uci}. Specifically, we consider Parkinson’s disease detection (D1), energy efficiency (D2), Istanbul stock exchange (D3), concrete slump test (D4), and QSAR aquatic toxicity (D5). To ensure overparameterization, we augment the feature space using the Nyström method~\citep{yang2012nystrom}. The details of experimental setup are discussed in the supplementary materials.

We compare CPCR against nine widely used latent factor–based methods, including dimension reduction techniques, regularized regression models, and nonparametric kernel approaches. Specifically, these baselines consist of PCR, ridge regression, Ledoit–Wolf OLS (LW-OLS)~\citep{ledoit2004well} Debiased Lasso (De-Lasso)~\citep{van2014asymptotically}, kernel regression (Kernel Reg)~\citep{ullah1999nonparametric}, kernel ridge regression (Kernel Ridge Reg)~\citep{murphy2012machine}, sliced inverse regression (SIR)~\citep{li1991sliced}, supervised PCA~\citep{bair2006prediction}. For every method, both the number of latent variables (when applicable) and the regularization parameters are selected independently using 5-fold cross-validation over each method’s own hyperparameter grid.

\begin{table*}[t]
        \centering
        \caption{RMSE and $R^2$ scores for 9 baselines compared against CPCR. CPCR consistently achieves better or competitive performance compared to the baselines with respect to both metrics. Mean values over $10$ repeated runs are reported. $\texttt{NV}$ stands for negative values.}
        \resizebox{1\textwidth}{!}{
        \begin{tabular}{llcccccccccc}
        \toprule
            &  & CPCR & PCR & Ridge & Ridgeless & LW-OLS & De-Lasso & Kernel Reg &Kernel Ridge & SIR & Super PCA \\
        \midrule
          \multirow{2}{*}{D1} & RMSE & \textbf{0.31} & 0.48 & \textbf{0.32} & 0.69 & \textbf{0.32} & 20.94& 0.34& 0.35& 0.73&0.37 \\
          & $R^2$ & \textbf{0.48}& $\texttt{NV}$ & 0.44 & $\texttt{NV}$ & 0.45 & $\texttt{NV}$& 0.26& 0.27& $\texttt{NV}$ &0.20\\
          \midrule
         \multirow{2}{*}{D2}  & RMSE & \textbf{0.16}&0.36 & 0.26 & 1.72 &2.22 & 2.18 &0.27 & 0.26& 6.99& 0.23\\
          & $R^2$ & \textbf{0.97}&0.86 & 0.93 & 0.96 &0.95 & $\texttt{NV}$ & 0.93&0.93 & $\texttt{NV}$ &0.95\\
          \midrule
          \multirow{2}{*}{D3} & RMSE & \textbf{0.31}&0.33 & 0.33 & 3.64 & 0.40&330.55 & 0.54&0.41 & 1.00&0.35\\
          & $R^2$ & \textbf{0.88}& 0.86& 0.86 & $\texttt{NV}$ & 0.79 &$\texttt{NV}$ & 0.69& 0.82& $\texttt{NV}$&0.87\\
          \midrule
          \multirow{2}{*}{D4}  & RMSE & \textbf{0.20}& 0.25 & \textbf{0.21} & 0.59 & \textbf{0.21}& 9.36& 0.27& 0.27& 0.32&0.28\\
          & $R^2$ & \textbf{0.52}& 0.30 & 0.49 & $\texttt{NV}$ & 0.50& $\texttt{NV}$ & 0.10&0.12  &$\texttt{NV}$ &0.02\\
          \midrule
          \multirow{2}{*}{D5}  & RMSE & \textbf{0.65}&3.99 & 0.66 & 0.96 & 1.24 & 170.29& 0.73& 0.64&4.93 &0.65\\
          & $R^2$ & \textbf{0.59}&$\texttt{NV}$ & 0.57 & 0.06 &0.45 &$\texttt{NV}$ & 0.37& 0.52& $\texttt{NV}$& 0.50\\
          \bottomrule
        \end{tabular}
        }
        \label{table:kernel_regression}
    \end{table*}

\paragraph{Results.} Table~\ref{table:kernel_regression} reports results evaluated by root-mean-square error (RMSE) and coefficient of determination ($R^2$). The results indicate that CPCR consistently achieves better or competitive performance compared to the benchmark algorithms. The reported results are mean values over 10 repeated runs. 

\subsection{Classification with Vision Foundation Model}
\paragraph{Dataset and methodology.}  
We evaluate CPCR on a high-dimensional classification task using the PACS dataset \citep{li2017pacs}, which consists of images from four distinct visual styles: Photo (P), Art Painting (A), Cartoon (C), and Sketch (S). Each image belongs to one of seven object categories (dog, elephant, giraffe, guitar, horse, house, person). Since the label depends on semantic content rather than artistic style, it is natural to expect that the predictive signal lies in a low-dimensional subspace of the high-dimensional feature space. 

We extract image embeddings using the latest DINOv3 vision foundation models~\citep{siméoni2025dinov3}. These models are based on ConvNeXt backbones~\citep{liu2022convnet} and produce feature vectors with dimensions ranging from 768 to 1536. To reflect the overparameterized regime, we restrict supervision to fewer than $500$ labeled examples for training and use the remaining $\sim$9500 labeled images for evaluation. To simulate  noise in the response, we randomly flip $20\%$ of the training labels. The subspace $\hU$ is learned from all available unlabeled features, with dimension fixed at $r=8$. 

\paragraph{Results.} Table~\ref{tab:pacs_model_comparison} reports test accuracy across logistic regression (LR), principal component regression (PCR), and our calibrated variant (CPCR). CPCR consistently improves upon both LR and PCR across all feature extractors, despite using only 8 principal components. These results highlight CPCR’s ability to mitigate PCR’s truncation bias and achieve stable gains in modern overparameterized vision settings.





\begin{table}[ht]
    \caption{
    Test accuracy on PACS dataset using ConvNeXt features from DINOv3. 
    CPCR (with $r=8$ PCs) consistently outperforms LR and PCR across model scales. 
    Mean $\pm$ standard deviation over 5 repeated runs is reported.
    }

    \centering
    \resizebox{1\linewidth}{!}{
    \begin{tabular}{lccc}
        \toprule
        \textbf{Model} & \textbf{LR} & \textbf{PCR} & \textbf{CPCR} \\
        \midrule
        ConvNext-tiny & 90.19 $\pm$ 0.40 & 86.88 $\pm$ 2.14 & \textbf{92.91 $\pm$ 1.78} \\
        ConvNext-small & 90.90 $\pm$ 1.13 & 86.96 $\pm$ 1.30 & \textbf{93.78 $\pm$ 0.76} \\
        ConvNext-base & 92.15 $\pm$ 0.56 & 90.33 $\pm$ 1.12 & \textbf{95.28 $\pm$ 0.38} \\
        ConvNext-large & 93.15 $\pm$ 0.67 & 92.33 $\pm$ 0.91 & \textbf{96.25 $\pm$ 0.42} \\
        \bottomrule
    \end{tabular}
    }
    \label{tab:pacs_model_comparison}
\end{table}


\section{DISCUSSION AND CONCLUSION}
We propose CPCR, a principled extension of PCR that leverages cross-fitting to mitigate truncation bias induced by hard spectral cutoff.
CPCR enjoys provable guarantees and achieves stable and robust improvements across a range of settings.


A limitation of our current analysis is that it focuses exclusively on in-distribution risk. In many practical scenarios, however, the test distribution may differ from the training distribution, as commonly encountered in domain generalization problems \citep{gulrajani2021search, zhu2026domain}. 
Extending the CPCR framework and its theoretical guarantees to settings with distribution shift, particularly under covariate or posterior shift, remains an important direction for future work and may further broaden its applicability.



\newpage

\subsubsection*{Acknowledgements}

The authors would like to thank Laura Balzano and Clayton Scott for helpful discussions. Authors acknowledge the support from the Department of Defense, Defense Threat Reduction Agency under award HDTRA1-20-2-0002, and the support from Northwestern University.



\bibliographystyle{plainnat}
\bibliography{reference}

\appendix


%
%


\renewcommand{\bibname}{References}
\renewcommand{\bibsection}{\subsubsection*{\bibname}}


%
\runningtitle{Calibrated Principal Component Regressions: Supplementary Materials}

%

\onecolumn
\aistatstitle{Calibrated Principal Component Regressions:  \\ Supplementary Materials}
\thispagestyle{empty}

This supplementary material is organized as follows. Section~\ref{sec1} restates key results from random matrix theory that are used throughout the proofs. Section~\ref{sec2} presents two auxiliary lemmas that support the main theoretical results, whose proofs are given in Section~\ref{sec3}. Finally, Section~\ref{sec4} provides additional experimental details.

\section{RANDOM MATRIX THEORY}\label{sec1}
Before delve into our proofs, we remind several related definitions and useful theorems from the random matrix theory in this section. The first very important tool is the Stieltjes transform, where we define as follow:

\begin{definition}[Stieltjes transform.] For a real probability measure $\mu$ with support $\text{supp}(\mu)$. For all $z\in\BC \setminus \text{supp}(\mu)$ the Stieltjes transform $m_\mu(z)$ is defined as 
\begin{align*}
    m_\mu(z) \equiv \int\frac{1}{t-z}\mu(dt) \, .
\end{align*}
\end{definition}

Knowing Stieltjes transform allows us to state one of the most important theorems in random matrix theory, which is the Marchenko–Pastur theorem. Here, we will introduce a slightly varied version, which is the generalized Marchenko–Pastur theorem. 

\paragraph{Notations.} In the following theorem and remaining parts of this supplementary material, we use the superscript $p$ to indicate the finite-$p$ companion Stieltjes transform $\Tilde{m}^p(z)$, and the subscript $z$ to indicate the derivative with respect to $z$, $\partial_z\Tilde{m}(z)=\Tilde{m}_z(z)$. $\Tilde{m}(z)$ is the limit of $\Tilde{m}^p(z)$ such that $n,p\rightarrow\infty$ with $p/n\rightarrow c \in (0,\infty)$. We denote $X \leftrightarrow Y$ if, for all bounded norm $\A\in \BR^{n \times n}$, and $\veca, \vecb \in \BR^n$, $\frac{1}{n}\tr\A(\X-\Y)\xrightarrow{a.s.}0$, $\veca^\top(\X-\Y)\vecb\xrightarrow{a.s.}0$, and $\norm{\BE[\X-\Y]}\rightarrow0$. Finally, we denote $\X_n \asymp \Y_n$ of matrices with growing dimensions if they are equivalent, that is $\lim_{n\rightarrow\infty}\frac{1}{n}\tr\A_n(\X_n-\Y_n)\xrightarrow{a.s.}0$ for any sequence of $\A_n$ with bounded norms.

\begin{theorem}[Generalized Marchenko–Pastur theorem,~\citet{silverstein1995empirical}]\label{thrm:MP} Let $\X=\bm{\Sigma}^{\frac{1}{2}}\Z \in \BR^{p \times n}$, where entries of $\Z$ have independent zero mean and unit variance, and light tail. $\bm{\Sigma}\BR^{p \times p}$ is a symmetric nonnegative definite population covariance matrix. Assume that as $n,p\rightarrow\infty$ with $p/n\rightarrow c \in (0,\infty)$. Letting $\bQ(z)=(\frac{1}{n}\X\X^\top-z\I_p)^{-1}$ and $\Tilde{\bQ}(z)(\frac{1}{n}\X^\top\X-z\I_n)^{-1}$, we have 
\begin{align} \label{eq:equivalent_mat}
    \bQ(z) &\leftrightarrow \bar{\bQ}(z) = -\frac{1}{z}(\I_p+\Tilde{m}^p(z)\bm{\Sigma})^{-1}, \\
    \Tilde{\bQ}(z)&\leftrightarrow \bar{ \Tilde{\bQ}}(z) = \Tilde{m}^p(z)\I_n\, , \nonumber
\end{align} 
where $(z,\Tilde{m}_p(z))$ is the unique solution in $\calZ(\BC  \setminus \BR^+)$ of 
\begin{align} \label{eq:tilde_m_FP}
    \Tilde{m}^p(z)=\left(-z + \frac{1}{n}\tr\bm{\Sigma}(\I+\Tilde{m}^p(z)\bm{\Sigma})^{-1} \right)^{-1} \, .
\end{align}
In particular, if the empirical spectral measure of $\bm{\Sigma}$ converges, i.e., $\mu_{\bm{\Sigma}}\rightarrow\nu$ as $p \rightarrow \infty$, then $\mu_{\frac{1}{n}\X\X^\top}\xrightarrow{a.s.} \mu$, $\mu_{\frac{1}{n}\X^\top\X}\xrightarrow{a.s.} \Tilde{\mu}$ as $n,p\rightarrow\infty$, where $\mu$ and $\Tilde{\mu}$ are the unique measure having Stieltjes transforms $m(z)$ and $\Tilde{m}(z)$, respectively, with the forms
\begin{align}\label{eq:tilde_m_limit}
    &m(z)=\frac{1}{c}\Tilde{m}(z)+\frac{1-c}{cz}\, , \nonumber \\
    &\Tilde{m}(z)=\left( -z+c\int\frac{t\nu(dt)}{1+\Tilde{m}(z)t} \right)^{-1} \, .
\end{align}

\end{theorem}


    

\newpage
 
\section{ADDITIONAL LEMMAS}\label{sec2}
In this section, we introduce two auxiliary lemmas to support the proof of Theorem~\ref{thrm:1-sup}. These two technical lemmas allow us to deal with the correlation between $\vgdebias_1{}$ and $\vgdebias_2{}$. 

\begin{lemma}[Vanishing direction of estimator.]\label{lemma:1}
    Under the same settings as Theorem~\ref{thrm:1-sup}, denote $\bsigmaone=\frac{1}{n_1}\X_1\X_1^\top$ the sample covariance of one random split of $\X$. Then, 
    \begin{align*}
        \U(\U^\top\bsigmaone\U)^{-1}\U^\top \bsigmaone  (\bm{\Pi}_{\U}-\I)\bm{\gamma}^*  \xrightarrow{a.s.} \mathbf{0}\, .
    \end{align*}
\end{lemma}

\begin{proof}
    Recall from \citet[Ch.~2, Sec.~2.4.2, p.109]{couillet2022random}
\begin{align*}
    \veca^\top f\left(\frac{1}{n}\X\X^T \right)\vecb = \frac{1}{2\pi i}\oint\limits_{\Gamma} \frac{f(z)}{z}\veca^\top (\I+\Tilde{m}(z)\bm{\Sigma})^{-1}\vecb \,dz + o(1)\,,
\end{align*}
where $f: \mathbb{C} \rightarrow \mathbb{C}$ analytic on the neighborhood of eigenvalue of $\frac{1}{n}\X\X^T$, $\Gamma$ is a contour circling around the limiting spectral support $\text{supp}(\mu)$ ($\mu$ as in Theorem~\ref{thrm:MP}), and $\bm{\Sigma}$ is the population covariance of $\X$. Here, if we take $\veca=\U$, $\vecb=\I-\bm{\Pi}_{\U}$, and $f(z)=z$, we have 
\begin{align*}
    \U^\top \bsigmaone (\I-\bm{\Pi}_{\U}) =& \frac{1}{2\pi i}\oint\limits_{\Gamma} \U^\top (\I+\Tilde{m}(z)\bm{\Sigma})^{-1}(\I-\bm{\Pi}_{\U}) \,dz + o(1) \\
    =& \frac{1}{2\pi i}\oint\limits_{\Gamma} \U^\top ( \U(I+\Tilde{m}(z)\bm{\Sigma}_s)^{-1}\U^\top + \V(I+\Tilde{m}(z)\bm{\Sigma}_c)^{-1}\V^\top  ) (\I-\bm{\Pi}_{\U}) \,dz + o(1) .
\end{align*}
Now, because $\U^\top (\I-\bm{\Pi}_{\U}) = \mathbf{0}$ and $\U^\top \V=\mathbf{0}$, the integrand is $\mathbf{0}$. Therefore, 
$$\hat{\U}(\hat{\U}^\top\bsigmaone\hat{\U})^{-1}\hat{\U}^\top \bsigmaone  (\bm{\Pi}_{\U}-\I)\bm{\gamma}^*  \xrightarrow{a.s.} \mathbf{0}\, .$$

\end{proof}

\begin{lemma}[Vanishing variance.]\label{lemma:2}
    Under the same settings as Theorem~\ref{thrm:1-sup}, denote $\bsigmatwo=\frac{1}{n_2}\X_2\X_2^\top$ the sample covariance of one random split of $\X$. Then, with high probability, 
    \begin{align*}
         \frac{\lambda^2\sigma^2}{n_2}\tr((\bsigmatwo+\lambda\I)^{-1}\Sigma (\bsigmatwo+\lambda\I)^{-1} \U(\U^\top\bsigmaone\U)^{-1}\U^\top ) \leq \calO(\frac{r}{n})
    \end{align*}
\end{lemma}

\begin{proof}
    We will tackle by bounding $(\U^\top\bsigmaone\U)^{-1}$ and $\U^\top(\bsigmatwo+\lambda\I)^{-1}\Sigma (\bsigmatwo+\lambda\I)^{-1} \U$.

Write $\U^\top\bsigmaone\U = \frac{1}{n_1}\U^\top \X_1 \X_1^\top \U$. Then $\bm{\Sigma}_{\U}=\BE(\frac{1}{n_1}\U^\top \X_1 \X_1^\top \U)=\U^\top\bm{\Sigma}\U=\bm{\Sigma}_s$ (Because $\BE(\bsigmaone)=\frac{1}{n_1}\sum_{i=1}^{n_1}\BE(\vecx_i\vecx_i^\top)=\BE(\vecx_1\vecx_1^\top)=\bm{\Sigma}$). Define $\W=\bm{\Sigma}_{\U}^{-1/2}\U^\top \X_1$, and write $\frac{1}{n_1}\U^\top \X_1 \X_1^\top \U = \Sigma_{\U}^{1/2}\frac{1}{n_1}\W\W^\top \bm{\Sigma}_{\U}^{1/2}$. Therefore, 
\begin{align*}
    \norm{(\U^\top\bsigmaone\U)^{-1}} = \norm{(\bm{\Sigma}_{\U}^{1/2}\frac{1}{n_1}\W\W^\top \bm{\Sigma}_{\U}^{1/2})^{-1}} &\leq \norm{\bm{\Sigma}_{\U}^{-1}}\norm{\left(\frac{1}{n_1}\W\W^\top \right)^{-1}}  \\
    &= \frac{1}{\lambda_{\min}(\bm{\Sigma}_{\U})}\frac{1}{\lambda_{\min}\left(\frac{1}{n_1}\W\W^\top \right)}
\end{align*}
by the multidisciplinary of matrix norm. Note that $\lambda_{\min}(\bm{\Sigma}_{\U})=\lambda_{\min}(\bm{\Sigma}_s)$. Also note that each column $\vecw$ of $\W=\bm{\Sigma}_{\U}^{-1/2}\U^\top\Hmat^{1/2}\Z_1$ has covariance $\BE(\vecw \vecw^\top) = \bm{\Sigma}_{\U}^{-1/2}\U^\top\Hmat \U \bm{\Sigma}_{\U}^{-1/2} = \I$ and mean $\mathbf{0}$. From \citet{Davidson-Szarek-2003} Theorem II.13, 
\begin{align*}
    \BP\left( \lambda_{\min}\left(\frac{1}{n_1}\W\W^\top \right) > (1-\sqrt{r/n_1} -t)^2 \right) \geq 1-\exp \left( \frac{-n_1t^2}{2} \right) \, .
\end{align*}

So, with high probability (choose $t=n_1^{-1/4}$), 
\begin{align*}
    \norm{(\U^\top\bsigmaone\U)^{-1}} \leq \frac{1}{\lambda_{\min}(\bm{\Sigma}_s)} \frac{1}{(1-\sqrt{r/n_1} -t)^2} \, .
\end{align*}

Next, we bound $\U^\top(\bsigmatwo+\lambda\I)^{-1}\Sigma (\bsigmatwo+\lambda\I)^{-1} \U$:
\begin{align*}
    \norm{\U^\top(\bsigmatwo+\lambda\I)^{-1}\Sigma (\bsigmatwo+\lambda\I)^{-1} \U} &\leq \norm{(\bsigmatwo+\lambda\I)^{-1}}^2 \norm{\bm{\Sigma}} \\
    &= \frac{\norm{\bm{\Sigma}}}{\lambda_{\min} (\bsigmatwo+\lambda\I)^2} \leq \frac{\norm{\bm{\Sigma}} }{\lambda^2}
\end{align*}

Put together, 
\begin{align*}
    \tr((\bsigmatwo+\lambda\I)^{-1}\Sigma (\bsigmatwo+\lambda\I)^{-1} \U(\U^\top\bsigmaone\U)^{-1}\U^\top ) 
    &\leq |\tr(\U^\top(\bsigmatwo+\lambda\I)^{-1}\bm{\Sigma} (\bsigmatwo+\lambda\I)^{-1} \U)| \norm{(\U^\top\bsigmaone\U)^{-1}} \\
    &\leq r\cdot \norm{\U^\top(\bsigmatwo+\lambda\I)^{-1}\bm{\Sigma} (\bsigmatwo+\lambda\I)^{-1} \U}\norm{(\U^\top\bsigmaone\U)^{-1}} \\
    &\leq r \cdot \frac{\norm{\bm{\Sigma}} }{\lambda^2} \frac{1}{\lambda_{\min}(\bm{\Sigma}_s)}\frac{1}{(1-\sqrt{r/n_1} -t)^2}  \, ,
\end{align*}
where the first inequality can be obtained either from Holder's inequality or von Neumann's trace inequality. The second inequality is from $|\tr(\A)|=|\sum_{i}\lambda_i(\A)| \leq r\cdot |\lambda_{\max}(\A)| = r \cdot \norm{\A}_{op}$ for $\A$ PSD. 

Note that $n_1=n_2=\frac{n}{2}$. Therefore, 
\begin{align*}
    \frac{\lambda^2\sigma^2}{n_2}\tr((\bsigmatwo+\lambda\I)^{-1}\Sigma (\bsigmatwo+\lambda\I)^{-1} \U(\U^\top\bsigmaone\U)^{-1}\U^\top ) &\leq \frac{r}{n} \cdot \frac{\norm{\bm{\Sigma}}}{\lambda_{\min}(\bm{\Sigma}_s)}\frac{2\sigma^2 }{(1-\sqrt{r/n_1} -t)^2}\\& =\calO\left(\frac{r}{n}\right) \,. 
\end{align*}
\end{proof}

\newpage
\section{MAIN THEOREMS}\label{sec3}
We will first restate our main theorems and prove them in the subsequent subsections. For the following calculations, recall that our out-of-sample prediction risk is defined as:
\begin{align}\label{eq:sup-risk}
    \mathcal{R}(\hat{\vgamma})\overset{\Delta}{=}\BE_{\vecx_0,\vgamma^*}[ (\vecx_0^\top \hat{\vgamma} -  \vecx_0^\top \vgamma^*)^2 ] \, .
\end{align}

\begin{theorem}[Exact calibration estimator risk.]\label{thrm:1-sup}
In the asymptotic limit where $n,p \rightarrow\infty$ such that $p/n \rightarrow c\in(1,\infty)$. With high probability, the risk of our calibration estimator $\hat{\bm{\gamma}}^{\mathtt{calib}}$ converge to a deterministic limit: 

\begin{align}\label{eq:sup-limit_theory}
    \mathcal{R}(\hat{\vgamma}^{\mathtt{CPCR}}) \rightarrow
    B_{\X}(\hat{\vgamma}^{\mathtt{CPCR}}) + V(\hat{\vgamma}^{\mathtt{CPCR}}) \, ,
\end{align}
where the bias term can be further estimated by
\begin{align*}
    B_{\X}(\hat{\vgamma}^{\mathtt{CPCR}}) = &\frac{1}{4}(B_{\X}(\hat{\bm{\gamma}}^{\mathtt{calib}}_1) + B_{\X}(\hat{\bm{\gamma}}^{\mathtt{calib}}_2)) \\+&\frac{1}{2}B_{\X}(\hat{\bm{\gamma}}^{\mathtt{calib}}_1, \hat{\bm{\gamma}}^{\mathtt{calib}}_2)
\end{align*}
and the variance term can be estimated by
\begin{align*}
     V(\hat{\vgamma}^{\mathtt{CPCR}}) = \frac{1}{4}(V_{\epsilon_2}(\hat{\bm{\gamma}}^{\mathtt{calib}}_1) + V_{\epsilon_1}(\hat{\bm{\gamma}}^{\mathtt{calib}}_2)) \, .
\end{align*}
Moreover, consider $\{\mu_j(\bm{\Sigma}_c)\}_{j=1}^{p-r}$ the diagonal entries of $\bm{\Sigma}_c$ and $\nu_c = \frac{1}{p-r}\sum_{j=1}^{p-r}\delta_{\mu_j(\bm{\Sigma}_c)}$, then the almost sure limits of each individual terms are
\begin{align}\label{eq:bias_1_limit-sup}
    &B_{\X}(\hat{\bm{\gamma}}^{\mathtt{calib}}_1) \xrightarrow{\text{a.s.}}\\ 
    +&(1-\kappa)(p-r)\int\frac{\mu}{1+\Tilde{m}(-\lambda)\mu} d\nu_c(\mu) \nonumber\\
    +&(1-\kappa)(p-r)\Tilde{m}_\rho(-\lambda) \int\frac{\mu}{(1+\Tilde{m}(-\lambda)\mu)^2} d\nu_c(\mu)\nonumber\\
    -& (1-\kappa)(p-r)\Tilde{m}(-\lambda)\int\frac{\mu^2}{(1+\Tilde{m}(-\lambda)\mu)^2} d\nu_c(\mu) \nonumber  \, ,
\end{align}
\begin{align}\label{eq:bias_1_2_limit-sup}
    &B_{\X}(\hat{\bm{\gamma}}^{\mathtt{calib}}_1, \hat{\bm{\gamma}}^{\mathtt{calib}}_2) \xrightarrow{\text{a.s.}} \\
    & (1-\kappa) (p-r)\int\frac{\mu}{(1+\Tilde{m}(-\lambda)\mu)^2} d\nu_c(\mu) \nonumber
\end{align} and
\begin{align} \label{eq:var_e2_limit-sup}
    V_{\epsilon_2}(\hat{\bm{\gamma}}^{\mathtt{calib}}_1)  \xrightarrow{\text{a.s.}} \sigma^2  \left( \frac{\Tilde{m}_z(-\lambda)}{\Tilde{m}^2(-\lambda)}-1 \right). 
\end{align}
$\Tilde{m}(\cdot)$ is defined implicitly as the solutions to a nonlinear equation depending on $n, p$ and $\bm{\Sigma}$. $\Tilde{m}_\rho(\cdot)$ and $\Tilde{m}_z(\cdot)$ are partial derivatives with respect to $\rho$ and $z$ respectively. 
\end{theorem} 

\begin{theorem}[Optimal $\lambda$.] \label{thrm:2-sup}
    Under the same conditions as Theorem~\ref{thrm:1-sup}, the optimal lambda $\lambda^*$ to achieve the minimal risk in~\eqref{eq:sup-risk} satisfies the following equation:
    \begin{align*}
        B_{\X}'(\lambda^*)+V'(\lambda^*) = 0 \, .
    \end{align*}
    In particular, 
    \begin{align*}
        B_{\X}'(\lambda) =& \frac{(1-\kappa)(p-r)}{2} \biggl( 
        2\int\frac{\mu^2\Tilde{m}_z(-\lambda)}{(1+\Tilde{m}(-\lambda)\mu)^2} d\nu_c(\mu) \\     
        &- \int\frac{\mu\Tilde{m}_{\rho z}(-\lambda) }{(1+\Tilde{m}(-\lambda)\mu)^2} d\nu_c(\mu) \\       
        &+ 2 \int\frac{\mu^2(\Tilde{m}_{\rho}(-\lambda)\Tilde{m}_z(-\lambda)+\Tilde{m}_z(-\lambda))}{(1+\Tilde{m}(-\lambda)\mu)^3} d\nu_c(\mu) \\       
        &-  2\int\frac{\mu^3\Tilde{m}(-\lambda)\Tilde{m}_z(-\lambda)}{(1+\Tilde{m}(-\lambda)\mu)^3} d\nu_c(\mu) 
        \biggl)
    \end{align*}
    
    and
    \begin{align*}
        V'(\lambda) = \frac{\sigma^2}{2} \biggl( \frac{-\Tilde{m}(-\lambda)\Tilde{m}_{zz}(-\lambda)+2\Tilde{m}_z^2(-\lambda)}{\Tilde{m}^3(-\lambda)} \biggl) \, .
    \end{align*}
    $\Tilde{m}_{\rho z}(\cdot)$ denotes $\partial_{\rho z}\Tilde{m}^p(\cdot)$ and $\Tilde{m}_{z z}(\cdot)$ denotes $\partial_{zz}\Tilde{m}^p(\cdot)$. 
\end{theorem}





\subsection{Proof of Theorem 1}
We will break the calculations into three calculations. First, we will give the deterministic equivalents for bias and variance of $\hat{\bm{\gamma}}^{\mathtt{calib}}_1$. Then we will give the deterministic equivalent for $B_{\X}(\hat{\bm{\gamma}}^{\mathtt{calib}}_1, \hat{\bm{\gamma}}^{\mathtt{calib}}_2)$. 


First, fitting a subspace regression model 
\begin{align*}
    \hat{\xi}=\arg \min_{\xi} \norm{\vecy_1^\top-\xi^\top\hat{\U}^\top\X_1}^2_2
\end{align*}
on $(\X_1, \vecy_1)$ gives
\begin{align}\label{Eq:xi_hat}
    \hat{\xi_1}= \left( \hat{\U}^\top \frac{\X_1\X^\top_1}{n_1} \hat{\U} \right)^{-1} \hat{\U}^\top \frac{\X_1\vecy_1}{n_1} \, .
\end{align}

Then, we calibrate $\bm{\gamma}$ with $\bm{\gamma}^\mathtt{init}_1=\hat{\U}\hat{\xi_1}$ and fit
\begin{align*}
    \hat{\bm{\gamma}}^{\mathtt{calib}}_1  = \arg \min_{\bm{\gamma}} \norm{\vecy^\top_2-\bm{\gamma}^\top\X_2}^2_2 + \lambda\norm{\bm{\gamma}-\bm{\gamma}^{\mathtt{init}}_1}^2_2\, .
\end{align*}
Substituting $\vecy_1=\X_1^\top\bm{\gamma}^*+\bm{\epsilon}_1$ and $\vecy_2=\X_2^\top\bm{\gamma}^*+\bm{\epsilon}_2$, we have 
\begin{align}\label{eq:gamma_calib-gamma_star}
    \hat{\bm{\gamma}}^{\mathtt{calib}}_1 - \bm{\gamma}^* &= \left(\left( \frac{\X_2\X^\top_2}{n_2} + \frac{\lambda}{n_2} \I \right)^{-1}\frac{\X_2\X^\top_2}{n_2} -\I \right)\bm{\gamma}^* + \left( \frac{\X_2\X^\top_2}{n_2} +\frac{\lambda}{n_2}\I \right)^{-1}\left(\frac{\X_2\bm{\epsilon}_2}{n_2} +\frac{\lambda}{n_2} \bm{\gamma}^{\mathtt{init}}_1 \right) \nonumber\\
    &= \frac{\lambda}{n_2} (\bsigmatwo+\frac{\lambda}{n_2} \I)^{-1}(\bm{\gamma}^{\mathtt{init}}_1-\bm{\gamma}^*)+(\bsigmatwo+\frac{\lambda}{n_2} \I)^{-1} \frac{\X_2\bm{\epsilon}_2}{n_2} \, ,
\end{align}
where $\hat{\Sigma}_i = \frac{\X_i\X^\top_i}{n_i}$ for $i=1,2$. The second equality comes from $\left(\bsigmatwo + \frac{\lambda}{n_2} \I \right)^{-1}\bsigmatwo -\I = \left(\bsigmatwo + \frac{\lambda}{n_2} \I \right)^{-1} \left(\bsigmatwo+ \frac{\lambda}{n_2} \I - \frac{\lambda}{n_2} \I\right) -\I = -\frac{\lambda}{n_2} \left( \bsigmatwo + \frac{\lambda}{n_2} \I \right)^{-1} $. For notation simplicity, from here we rescale $\frac{\lambda}{n_2} =\frac{\lambda}{n_1}$ to $\lambda$. 

Next, substituting $\vecy_1$ into~\eqref{Eq:xi_hat} gives
\begin{align*}
    \hat{\xi}_1=(\hat{\U}^\top\bsigmaone\hat{\U})^{-1}\hat{\U}^\top\bsigmaone\bm{\gamma}^* + (\hat{\U}^\top\bsigmaone\hat{\U})^{-1}\hat{\U}^\top \frac{\X_1\bm{\epsilon}^\top_1}{n_1},
\end{align*}
and with $\bm{\gamma}^\mathtt{init}_1=\hat{\U}\hat{\xi_1}$ we have
\begin{align}\label{eq:gamma_init-gamma_star}
    \bm{\gamma}^{\mathtt{init}}_1-\bm{\gamma}^* = (\hat{\U}(\hat{\U}^\top\bsigmaone\hat{\U})^{-1}\hat{\U}^\top\bsigmaone-\I)\bm{\gamma}^* + \hat{\U}(\hat{\U}^\top\bsigmaone\hat{\U})^{-1}\hat{\U}^\top \frac{\X_1\bm{\epsilon}^\top_1}{n_1}.
\end{align}

Substituting~\eqref{eq:gamma_init-gamma_star} into~\eqref{eq:gamma_calib-gamma_star}, we have
\begin{align}
    &\hat{\bm{\gamma}}^{\mathtt{calib}}_1 - \bm{\gamma}^* \nonumber\\ =&  \lambda(\bsigmatwo+\lambda\I)^{-1}\left[ (\bm{\Pi}_{\U}-\I)\bm{\gamma}^* + \hat{\U}(\hat{\U}^\top\bsigmaone\hat{\U})^{-1}\hat{\U}^\top \frac{\X_1\epsilon_1^\top}{n_2} + \tred{ \hat{\U}(\hat{\U}^\top\bsigmaone\hat{\U})^{-1}\hat{\U}^\top \bsigmaone  (\bm{\Pi}_{\U}-\I)\bm{\gamma}^* } \right] + (\bsigmatwo+\lambda\I)^{-1}\frac{\X_2\epsilon_2^\top}{n_2} \,.\label{Eq:calib_estimator_diff}
\end{align}
By Lemma~\ref{lemma:1}, the \tred{red term} converges to $0$ almost surely. Therefore, the bias term is 
\begin{align}
    B_{\X}(\hat{\bm{\gamma}}^{\mathtt{calib}}_1) &= \lambda^2 \tr( \BE((\bm{\Pi}_{\U}-\I)\bm{\gamma}^*\bm{\gamma}^{*\top}(\bm{\Pi}_{\U}-\I)) \cdot  (\bsigmatwo+\lambda\I)^{-1}\Sigma(\bsigmatwo+\lambda\I)^{-1}) \nonumber \\
    &= \lambda^2 \tr(\BE(\bm{\gamma}^*_{\perp}\bm{\gamma}^{*\top}_{\perp})  \cdot  (\bsigmatwo+\lambda\I)^{-1}\Sigma(\bsigmatwo+\lambda\I)^{-1}) \nonumber \\
    &= \lambda^2 \tr(\Sigma_{\bm{\gamma}^*_{\perp}}  \cdot  (\bsigmatwo+\lambda\I)^{-1}\Sigma(\bsigmatwo+\lambda\I)^{-1})\, . \label{Eq:bias_1} 
\end{align}


Recall that $\bm{\Sigma}_{\bm{\gamma}^*}=\kappa\bm{\Pi}_{\U}+(1-\kappa)\bm{\Pi}_{\V}$, giving $\bm{\Sigma}_{\bm{\gamma}^*_{\perp}}=(\I-\bm{\Pi}_{\U})\bm{\Sigma}_{\bm{\gamma}^*}(\I-\bm{\Pi}_{\U})= (1-\kappa)\bm{\Pi}_{\V}$. 
\begin{align}
    B_{\X}(\hat{\bm{\gamma}}^{\mathtt{calib}}_1) &= \lambda^2(1-\kappa) \tr( \bm{\Pi}_{\V}  \cdot  (\bsigmatwo+\lambda\I)^{-1}\bm{\Sigma}(\bsigmatwo+\lambda\I)^{-1}) \label{Eq:bias_gamma_assumed}
\end{align}



Observe from~\eqref{Eq:bias_gamma_assumed} that the appearance of $(\bsigmatwo+\lambda\I)^{-1}\bm{\Sigma}(\bsigmatwo+\lambda\I)^{-1}$. We cannot directly replace both resolvents $(\bsigmatwo+\lambda\I)^{-1}$ with their deterministic equivalents because they are correlated. As a result, we use some derivative tricks following~\citet{Tibshirani2024lecture}. 

We have: 
\begin{align*}
    \lambda^2 (\bsigmatwo+\lambda\I)^{-1}\bm{\Sigma}(\bsigmatwo+\lambda\I)^{-1} = \left. -\frac{d}{d\rho} \Biggl\{ \lambda(\bsigmatwo+\lambda(\I+\rho \bm{\Sigma})^{-1}) \Biggr\} \right|_{\rho=0}, 
\end{align*}
where the matrix in the curly bracket can be rewritten as 
\begin{align*}
    \lambda(\bsigmatwo+\lambda(\I+\rho \bm{\Sigma})^{-1}) = (\I+\rho \bm{\Sigma})^{-1/2}\lambda(\hat{\bm{\Sigma}}_{\rho}+\lambda\I)^{-1}(\I+\rho \bm{\Sigma})^{-1/2},
\end{align*}
where we define $\hat{\bm{\Sigma}}_{\rho}=\bm{\Sigma}_{\rho}^{1/2}\frac{Z_2Z_2^\top}{n_2}\bm{\Sigma}_{\rho}^{1/2}$ and $\bm{\Sigma}_{\rho}=(\I+\rho \bm{\Sigma})^{-1/2}\bm{\Sigma}(\I+\rho \bm{\Sigma})^{-1/2}$. Applying generalized MP law, we have
\begin{align*}
    & \lambda^2\bm{\Pi}_{\V} (\bsigmatwo+\lambda\I)^{-1}\bm{\Sigma}(\bsigmatwo+\lambda\I)^{-1} \\
    =& \left. - \bm{\Pi}_{\V}\cdot \frac{d}{d\rho} \Biggl\{ (\I+\rho \bm{\Sigma})^{-1/2}\lambda(\hat{\bm{\Sigma}}_{\rho}+\lambda\I)^{-1}(\I+\rho \bm{\Sigma})^{-1/2} \Biggr\} \right|_{\rho=0}  \\ 
    =& \left. - \bm{\Pi}_{\V}\cdot \frac{d}{d\rho} \Biggl\{ (\I+\rho \bm{\Sigma})^{-1/2}\lambda \bQ(-\lambda;\rho)(\I+\rho \bm{\Sigma})^{-1/2} \Biggr\} \right|_{\rho=0}  \\
    \asymp& \left. - \bm{\Pi}_{\V}\cdot \frac{d}{d\rho} \Biggl\{ (\I+\rho \bm{\Sigma})^{-1/2}(\I + \Tilde{m}^p(-\lambda)\bm{\Sigma}_\rho)^{-1}(\I+\rho \bm{\Sigma})^{-1/2} \Biggr\} \right|_{\rho=0} 
\end{align*}
where $\bQ(z;\rho)$ is the resolvent for $\hat{{\bm{\Sigma}}}_\rho$ with its deterministic equivalence $\bar{\bQ}(z;\rho)$. 

Now we proceed to derivative calculations. For simplicity and future reference, we denote $\Tilde{m}_\rho(\cdot)$ and $\Tilde{m}_z(\cdot)$ as partial derivatives of $\Tilde{m}^p(z;\rho)$ with respect to $\rho$ and $z$ respectively. 
\begin{align*}
     &\frac{d}{d\rho} \Biggl\{ (\I+\rho \bm{\Sigma})^{-1/2}(\I+\Tilde{m}^p(-\lambda;\rho)\bm{\Sigma}_\rho)^{-1}(\I+\rho \bm{\Sigma})^{-1/2} \Biggr\} \\
     = -&\frac{1}{2}(\I+\rho \bm{\Sigma})^{-3/2}\bm{\Sigma} (\I+\Tilde{m}^p(-\lambda;\rho)\bm{\Sigma}_\rho)^{-1}(\I+\rho \bm{\Sigma})^{-1/2} -\frac{1}{2}(\I+\rho \bm{\Sigma})^{-1/2}(\I+\Tilde{m}^p(-\lambda;\rho)\bm{\Sigma}_\rho)^{-1}(\I+\rho \bm{\Sigma})^{-3/2}\bm{\Sigma} \\
     -&  (\I+\rho \bm{\Sigma})^{-1/2}(\I+\Tilde{m}^p(-\lambda;\rho)\bm{\Sigma}_\rho)^{-1}(\Tilde{m}_\rho(-\lambda;\rho)\bm{\Sigma}_\rho+\Tilde{m}^p(-\lambda;\rho)\bm{\Sigma}'_\rho)(\I+\Tilde{m}^p(-\lambda;\rho)\bm{\Sigma}_\rho)^{-1}(\I+\rho \bm{\Sigma})^{-1/2} \, ,
\end{align*}
where we recall that $\frac{d\bm{A}^{-1}}{d\rho}=-\bm{A}^{-1}\frac{d\bm{A}}{d\rho}\bm{A}^{-1}$. 

We now calculate $\Tilde{m}_\rho(z;\rho)$ and $\bm{\Sigma}'_\rho$:
\begin{align*}
    &-\frac{\Tilde{m}_\rho(z;\rho)}{\Tilde{m}^2(z;\rho)}\\ =& \frac{1}{n_2}\tr(\bm{\Sigma}'_\rho(\I+\Tilde{m}^p(z;\rho)\bm{\Sigma}_\rho)^{-1})-\frac{1}{n_2}\tr(\bm{\Sigma}_\rho(\I+\Tilde{m}^p(z;\rho)\bm{\Sigma}_\rho)^{-1}(\Tilde{m}_\rho(z;\rho)\bm{\Sigma}_\rho+\Tilde{m}^p(z;\rho)\bm{\Sigma}'_\rho)(\I+\Tilde{m}^p(z)\bm{\Sigma}_\rho)^{-1})
\end{align*}

Therefore, $\Tilde{m}_\rho(z;\rho)$ has the closed form
\begin{align}
    &\Tilde{m}_\rho(z;\rho)\left(-\frac{1}{(\Tilde{m}^p(z))^2}+\frac{1}{n_2}\tr(\bm{\Sigma}_\rho(\I+\Tilde{m}^p(z;\rho)\bm{\Sigma}_\rho)^{-1}\bm{\Sigma}_\rho(\I+\Tilde{m}^p(z;\rho)\bm{\Sigma}_\rho)^{-1}) \right) \nonumber \\=& \frac{1}{n_2}\tr(\bm{\Sigma}'_\rho(\I+\Tilde{m}^p(z;\rho)\bm{\Sigma}_\rho)^{-1}) - \frac{1}{n_2}\tr(\bm{\Sigma}_\rho(\I+\Tilde{m}^p(z;\rho)\bm{\Sigma}_\rho)^{-1}\Tilde{m}^p(z;\rho)\bm{\Sigma}'_\rho(\I+\Tilde{m}^p(z;\rho)\bm{\Sigma}_\rho)^{-1})\, , \label{Eq:m'}
\end{align}
and the derivative of $\bm{\Sigma}_\rho$ reads
\begin{align*}
    \bm{\Sigma}'_\rho = -\frac{1}{2}(\I+\rho\bm{\Sigma})^{-3/2}\bm{\Sigma}^2(\I+\rho \bm{\Sigma})^{-1/2}-\frac{1}{2}(\I+\rho \bm{\Sigma})^{-1/2}\bm{\Sigma}(\I+\rho \bm{\Sigma})^{-3/2}\bm{\Sigma}\, .
\end{align*}

We simplify equations at $\rho=0$ and consider $z=-\lambda$: 
\begin{align*}
    &\left. \frac{d}{d\rho} \Biggl\{ \lambda(\bsigmatwo+\lambda(\I+\rho \bm{\Sigma})^{-1}) \Biggr\} \right|_{\rho=0} \\
    =& -\frac{1}{2}\bm{\Sigma}(\I+\Tilde{m}^p(-\lambda)\bm{\Sigma})^{-1}-\frac{1}{2}(\I+\Tilde{m}^p(-\lambda)\bm{\Sigma})^{-1}\bm{\Sigma}-(\I+\Tilde{m}^p(-\lambda)\bm{\Sigma})^{-1}(\Tilde{m}^p(-\lambda)\bm{\Sigma}-\Tilde{m}^p(-\lambda)\bm{\Sigma}^2)(\I+\Tilde{m}^p(-\lambda)\bm{\Sigma})^{-1}
\end{align*}
with
\begin{align*}
    \Tilde{m}^p(-\lambda) = \left(\lambda + \frac{1}{n_2}\tr(\bm{\Sigma}(\I+\Tilde{m}^p(-\lambda)\bm{\Sigma})^{-1}) \right)^{-1}
\end{align*}
and 
\begin{align*}
    &\Tilde{m}_\rho(-\lambda)\left(-\frac{1}{(\Tilde{m}^p(-\lambda))^2}+\frac{1}{n_2}\tr(\bm{\Sigma}(\I+\Tilde{m}^p(-\lambda)\bm{\Sigma})^{-1}\bm{\Sigma}(\I+\Tilde{m}^p(-\lambda)\bm{\Sigma})^{-1})\right) \\ =& -\frac{1}{n_2}\tr(\bm{\Sigma}^2(\I+\Tilde{m}^p(-\lambda)\bm{\Sigma})^{-1}) + \frac{1}{n_2}\tr(\bm{\Sigma}(\I+\Tilde{m}^p(-\lambda)\bm{\Sigma})^{-1}\Tilde{m}^p(-\lambda)\bm{\Sigma}^2(\I+\Tilde{m}^p(-\lambda)\bm{\Sigma})^{-1})\, ,
\end{align*}
Substitute back to~\eqref{Eq:bias_gamma_assumed} and we have the asymptotic limit of the bias:
\begin{align*}
    &B_{\X}(\hat{\bm{\gamma}}^{\mathtt{calib}}_1) \xrightarrow{\text{a.s.}} \\ 
    +&\frac{1-\kappa}{2}\tr(\bm{\Pi}_{\V}\bm{\Sigma}(\I+\Tilde{m}^p(-\lambda)\bm{\Sigma})^{-1}) \\
    +&\frac{1-\kappa}{2}\tr(\bm{\Pi}_{\V}(\I+\Tilde{m}^p(-\lambda)\bm{\Sigma})^{-1}\bm{\Sigma}) \\
    +&(1-\kappa)\tr(\bm{\Pi}_{\V}(\I+\Tilde{m}^p(-\lambda)\bm{\Sigma})^{-1}(\Tilde{m}_\rho(-\lambda)\bm{\Sigma}-\Tilde{m}^p(-\lambda)\bm{\Sigma}^2)(\I+\Tilde{m}^p(-\lambda)\bm{\Sigma})^{-1})\, .
\end{align*}
Further notice that projection onto the column space of $\V$ rules out the effects of $\bm{\Sigma}_s$ on the bias, so we could further simplify the bias term to:
\begin{align*}
    &B_{\X}(\hat{\bm{\gamma}}^{\mathtt{calib}}_1) \xrightarrow{\text{a.s.}} \\ 
    +&\frac{1-\kappa}{2}\tr(\bm{\Sigma}_c(\I+\Tilde{m}(-\lambda)\bm{\Sigma}_c)^{-1}) \\
    +&\frac{1-\kappa}{2}\tr((\I+\Tilde{m}(-\lambda)\bm{\Sigma}_c)^{-1}\bm{\Sigma}_c) \\
    +&(1-\kappa)\tr((\I+\Tilde{m}(-\lambda)\bm{\Sigma}_c)^{-1}(\Tilde{m}_\rho(-\lambda)\bm{\Sigma}_c-\Tilde{m}(-\lambda)\bm{\Sigma}_c^2)(\I+\Tilde{m}(-\lambda)\bm{\Sigma}_c)^{-1})\, .
\end{align*}

This concludes the first part of the proof. Note that in the main theorem we write trace as integral forms. 






There are two variance term related to $\hat{\bm{\gamma}}^{calib}_1$. The first term is the variance contributed by $\epsilon_1$:
\begin{align*}
    V_{\epsilon_1}(\hat{\bm{\gamma}}^{calib}_1) = \frac{\lambda^2\sigma^2}{n_2}\tr((\bsigmatwo+\lambda\I)^{-1}\bm{\Sigma} (\bsigmatwo+\lambda\I)^{-1} \hat{\U}(\hat{\U}^\top\bsigmaone\hat{\U})^{-1}\hat{\U}^\top )\, ,
\end{align*}
which we can bound using Lemma~\ref{lemma:2}. We also have variance contributed by $\epsilon_2$:
\begin{align}
    V_{\epsilon_2}(\hat{\bm{\gamma}}^{calib}_1) =& \frac{\sigma^2}{n_2}\tr(\bsigmatwo(\bsigmatwo+\lambda\I)^{-2}\bm{\Sigma}) \nonumber \\
    =&\frac{\sigma^2 p}{n_2}\left( \frac{1}{p}\tr(\bm{\Sigma}(\bsigmatwo+\lambda\I)^{-1}) - \frac{\lambda}{p}\tr(\bm{\Sigma}(\bsigmatwo+\lambda\I)^{-2}) \right) \label{eq:V_2}
\end{align}
We will again use generalized MP law to find the asymptotic deterministic limit of this variance term. By deterministic equivalence, we know that $\frac{1}{p}\tr(\bm{\Sigma}(\bsigmatwo+\lambda\I)^{-1})$ and $\frac{1}{p}\tr(\bm{\Sigma}\bar{\bQ}(-\lambda))=\frac{1}{p}\tr(\bm{\Sigma}(\lambda\Tilde{m}^p(-\lambda)\bm{\Sigma}+\lambda\I)^{-1})$ have the same asymptotic limit. To figure out $\Tilde{m}^p(-\lambda)$ in~\eqref{eq:equivalent_mat}, we rewrite~\eqref{eq:tilde_m_FP} and~\eqref{eq:tilde_m_limit} as

\begin{align}\label{eq:m_limit_rewrite_FP}
    \frac{1}{z\Tilde{m}^p(z)}+1= 2c \cdot \frac{1}{p}\tr(\bm{\Sigma}(z\I+z\Tilde{m}^p(z)\bm{\Sigma})^{-1})
\end{align}
and 
\begin{align}
    \frac{1}{z\Tilde{m}(z)}+1=2c\int\frac{t\nu(dt)}{z+z\Tilde{m}(z)t} \, . \label{eq:m_limit_rewrite}
\end{align}
By observing~\eqref{eq:m_limit_rewrite_FP} and~\eqref{eq:m_limit_rewrite}, we arrive at
\begin{align}\label{eq:v2_first_term}
    \frac{1}{p}\tr(\bm{\Sigma}(\bsigmatwo+\lambda\I)^{-1}) \xrightarrow{a.s.}\frac{1}{2c}\left(\frac{1}{\lambda\Tilde{m}(-\lambda)}-1\right) \,. 
\end{align}
We will apply the derivative trick to deal with $\frac{1}{p}\tr(\bm{\Sigma}(\bsigmatwo+\lambda\I)^{-2})$. Note that,
\begin{align*}
    \frac{1}{p}\tr(\bm{\Sigma}(\bsigmatwo+\lambda\I)^{-2}) =  -\frac{1}{p}\frac{d}{d\lambda} \Biggl\{ \tr(\bm{\Sigma}(\bsigmatwo+\lambda\I)^{-1})  \Biggr\} \, , 
\end{align*}
as a result, 
\begin{align}\label{eq:v2_second_term}
    \frac{1}{p}\tr(\bm{\Sigma}(\bsigmatwo+\lambda\I)^{-2}) \xrightarrow{a.s.}-\frac{d}{d\lambda} \Biggl\{ \frac{1}{2c}\left(\frac{1}{\lambda\Tilde{m}(-\lambda)}-1\right) \Biggr\} \, .
\end{align}
Substituting~\eqref{eq:v2_first_term} and~\eqref{eq:v2_second_term} into~\eqref{eq:V_2}, with 
\begin{align*}
    \Tilde{m}_z(-\lambda)\left(\frac{1}{\Tilde{m}^2(-\lambda)} -\frac{1}{n_2}\tr(\bm{\Sigma}(\I+\Tilde{m}(-\lambda)\bm{\Sigma})^{-1}\bm{\Sigma}(\I+\Tilde{m}(-\lambda)\bm{\Sigma})^{-1}) \right) =1 \, ,
\end{align*}
we have~\eqref{eq:var_e2_limit} and conclude the second part of our proof.

The remaining task is to identify the limit of the cross-bias term $B_{\X}(\hat{\bm{\gamma}}^{\mathtt{calib}}_1, \hat{\bm{\gamma}}^{\mathtt{calib}}_2)$. By applying the same steps as in~\eqref{eq:gamma_calib-gamma_star}-\eqref{Eq:calib_estimator_diff} and substituting into~\eqref{eq:sup-risk} for $\hat{\bm{\gamma}}^{\mathtt{CPCR}} = \frac{\hat{\bm{\gamma}}^{\mathtt{calib}}_1+\hat{\bm{\gamma}}^{\mathtt{calib}}_2}{2}$, we will have some risks terms for $\hat{\bm{\gamma}}^{\mathtt{calib}}_1$ and $\hat{\bm{\gamma}}^{\mathtt{calib}}_2$, respectively, and also the cross-bias term $B_{\X}(\hat{\bm{\gamma}}^{\mathtt{calib}}_1, \hat{\bm{\gamma}}^{\mathtt{calib}}_2)$ (It is easy to see the cross-variance terms vanish due to independence). More specifically, 
\begin{align*}
    B_{\X}(\hat{\bm{\gamma}}^{\mathtt{calib}}_1, \hat{\bm{\gamma}}^{\mathtt{calib}}_2)=\frac{1}{2}\lambda^2(1-\kappa)\tr( \bm{\Pi}_{\V}(\bsigmatwo+\lambda\I)^{-1} \bm{\Sigma} (\bsigmaone+\lambda\I)^{-1})\, .
\end{align*}
According to how we split the data, $\bsigmaone$ and $\bsigmatwo$ are independent. We treat $(\bsigmaone+\lambda\I)^{-1}$ as a deterministic matrix when applying deterministic equivalent for $(\bsigmatwo+\lambda\I)^{-1}$, and repeat for $(\bsigmaone+\lambda\I)^{-1}$. Finally, 
\begin{align*}
    B_{\X}(\hat{\bm{\gamma}}^{\mathtt{calib}}_1, \hat{\bm{\gamma}}^{\mathtt{calib}}_2) \xrightarrow{a.s.}\frac{1}{2}(1-\kappa)\tr((\I+\Tilde{m}(-\lambda)\bm{\Sigma}_c)^{-2}\bm{\Sigma}_c)\, .
\end{align*}
Again in the main theorem we represent trace using integral, and this concludes our proof for Theorem~\ref{thrm:1-sup}. 

\subsection{Proof of Theorem 2: Optimal $\lambda$}
The minimum risk is achieved when $\frac{d}{d\lambda}\calR=0$ with the corresponding optimal $\lambda^*$. It is straightforward to obtain $B_{\X}'(\lambda)$ and $V'(\lambda)$ stated in Theorem~\ref{thrm:2-sup}. Here, we just provide the explicit forms of $\Tilde{m}_{\rho z}(\cdot)$ and $\Tilde{m}_{z z}(\cdot)$. Recall from the previous proof:
\begin{align*}
    \Tilde{m}_z(-\lambda)\left(\frac{1}{\Tilde{m}^2(-\lambda)} -\frac{1}{n_2}\tr(\bm{\Sigma}(\I+\Tilde{m}(-\lambda)\bm{\Sigma})^{-1}\bm{\Sigma}(\I+\Tilde{m}(-\lambda)\bm{\Sigma})^{-1}) \right) =1 \, ,
\end{align*}
and denote $G(-\lambda)=\frac{1}{\Tilde{m}^2(-\lambda)} -\frac{1}{n_2}\tr(\bm{\Sigma}\A^{-1}\bm{\Sigma}\A^{-1})$ and $\A=\I+\Tilde{m}(-\lambda)\bm{\Sigma}$. Therefore, 
\begin{align*}
    \Tilde{m}_{zz}(-\lambda) = \frac{G'(-\lambda)}{G^3(-\lambda)} \, ,
\end{align*}
with $G'(-\lambda)=\frac{2}{\Tilde{m}^3(-\lambda)}-\frac{2}{n_2}\tr(\bm{\Sigma}\A^{-1}\bm{\Sigma}\A^{-1}\bm{\Sigma}\A^{-1})$.

Also recall:
\begin{align*}
    &\Tilde{m}_\rho(-\lambda)\left(-\frac{1}{\Tilde{m}^2(-\lambda)}+\frac{1}{n_2}\tr(\bm{\Sigma}(\I+\Tilde{m}(-\lambda)\bm{\Sigma})^{-1}\bm{\Sigma}(\I+\Tilde{m}(-\lambda)\bm{\Sigma})^{-1})\right) \\ =& -\frac{1}{n_2}\tr(\bm{\Sigma}^2(\I+\Tilde{m}(-\lambda)\bm{\Sigma})^{-1}) + \frac{1}{n_2}\tr(\bm{\Sigma}(\I+\Tilde{m}(-\lambda)\bm{\Sigma})^{-1}\Tilde{m}(-\lambda)\bm{\Sigma}^2(\I+\Tilde{m}(-\lambda)\bm{\Sigma})^{-1})\, ,
\end{align*}
and denote $H(-\lambda)=-\frac{1}{n_2}\tr(\bm{\Sigma}^2\A^{-1}) + \frac{\Tilde{m}(-\lambda)}{n}\tr(\bm{\Sigma}\A^{-1}\bm{\Sigma}^2\A^{-1})$. Therefore, 
\begin{align*}
    \Tilde{m}_{\rho z}(-\lambda) = \frac{GH'-HG'}{G^3(-\lambda)}\, ,
\end{align*}
with $H'(-\lambda)=-\frac{2}{n_2}\tr(\bm{\Sigma}^2\A^{-1}\bm{\Sigma}\A^{-1})+\frac{2\Tilde{m}(-\lambda)}{n_2}\tr(\bm{\Sigma}^2\A^{-1}\bm{\Sigma}\A^{-1}\bm{\Sigma}\A^{-1})$.

\newpage
\section[ADDITIONAL EXPERIMENT DETAILS]{ADDITIONAL EXPERIMENT DETAILS\protect\footnote{All code used in this work is available at \url{https://github.com/florencewuyixuan/CPCR}.}}\label{sec4}

\subsection{Kernel Regression with Nystrom Features}

In this section, we elaborate on the settings for the numerical experiments in 
\Cref{subsec:exp_nystrom}
that are omitted in the main paper. We choose $5$ datasets from the UCI Machine Learning repository~\citep{asuncion2007uci}:
\begin{enumerate}
    
\item D1: Parkinson's disease detection dataset~\citep{little2008suitability} that predicts Parkinson's disease. The input variables are the voice measurement from subjects, and the output is the status indicator about the disease.
\item D2: Energy efficiency dataset~\citep{tsanas2012accurate} that predicts the energy performance of residential building. The input variables are the features of building like orientation and relative compactness, and the response value is the energy load of the building.
\item D3: Istanbul stock exchange datasets ~\citep{akbilgic2014novel} that predicts the index returns. Similar to the origin paper, we adopt the ISE100 index as the response variable and the rest of the stocks market as input variables.
\item D4: Concrete slump test~\citep{concrete_slump_test_182} that predicts slump flow phenomenon. In this regression task, we choose compressive strength as response variable, and the features like cement and water as input variables.
\item D5: QSAR aquatic toxicity dataset~\citep{cassotti2014prediction} that predicts the  aquatic toxicity of different organic molecules. The input variables are the molecular descriptors, and the output is the toxicity.
\end{enumerate}

We use the Nystrom method \citep{yang2012nystrom} to construct nonlinear features based on raw input features. 
More specifically, for raw features on two input samples $x_i$ and $x_j$, we calculate the radial basis function (RBF) as $\kappa ( x _ { i } , x _ { j } ) = \text{exp} ( - | | x _ { i } - x _ { j } | | ^ { 2 } / ( 2 \sigma ^ { 2 } ) )$, where $\sigma$ is a length-scale parameter. The kernel Gram matrix $\widehat { K } = \left[ \kappa ( \widehat { x } _ { i } , \widehat { x } _ { j } ) \right] _ { m \times m }$ is then constructed by calculating the RBF function on a random subset of $m$ samples. Then, we adopt the Nystrom approximation as  $\widehat{K}_r=K_{b}\widehat{K}^{+}K_{b}^{T}$, in which  $ K _ { b } = \left[ \kappa ( x _ { i } , \widehat { x } _ { j } ) \right] _ { N \times m }  $, and $ \widehat { K }^{+} $ is the pseudo inverse of the kernel Gram matrix $\widehat{K}$. Cholesky decomposition on $\widehat{K}_r$ would yield the Nystrom features used in regression analysis. The number of instances, raw feature dimensions, and Nystrom feature dimensions of datasets are summarized in Table \ref{tab:dataset}.

\begin{table}[htbp]
\centering
\begin{tabular}{cccc}
\hline
   & $\#$ Raw Features & $\#$ Instances & $\#$ Nystrom Features \\ \hline
D1 & 22       & 197       & 200                \\
D2 & 8        & 768       & 800                \\
D3 & 9        & 536       & 600                \\
D4 & 10       & 103       & 200                \\
D5 & 8        & 546       & 600                \\ \hline
\end{tabular}
\caption{Summary of dataset characteristics}
\label{tab:dataset}
\end{table}



\subsection{Classification with Vision Foundation Model (Additional Details)}

\paragraph{Data preprocessing and feature extraction.}
We conduct experiments on the PACS benchmark, which comprises four domains (Photo, Art Painting, Cartoon, Sketch) and seven semantic classes. 
For each domain, we use pre-extracted image representations obtained from DINOv3 ConvNeXt encoders. 
Depending on the backbone size, the feature dimension ranges from 768 to 1536; for example, ConvNeXt-tiny yields 768-dimensional embeddings and ConvNeXt-large yields 1536-dimensional embeddings.

\paragraph{Train/test protocol.}
To emulate an extreme overparameterized regime, we perform a stratified random split within each domain, retaining only $5\%$ of labeled samples for training and using the remaining $95\%$ for evaluation. 
Across all domains, this results in approximately $500$ training samples in total and nearly $10^4$ evaluation samples. 
All domains are pooled together during both training and evaluation; no domain is held out.

\paragraph{Label noise model.}
We inject synthetic label noise into the training set by corrupting a fixed fraction (20\%) of labels independently. 
For each selected sample, the ground-truth label is replaced by a uniformly sampled incorrect class label (i.e., drawn uniformly from the remaining classes). This corresponds to a uniform label noise model \citep{zhu2024label}.

\paragraph{Model training.}
All methods are implemented as linear classifiers on top of pre-extracted, frozen foundation model features. 
For logistic regression (LR), we use an $\ell_2$-regularized multinomial model with the \texttt{lbfgs} solver, no intercept term, and a maximum of 200 iterations. 
The regularization parameter $C$ is selected from the grid 
$\{10^{-10}, 10^{-2}, 10^{-1}, 1, 10, 100, 10^{10}\}$.

For PCR and CPCR, we first compute principal components via singular value decomposition. 
Importantly, the PCA subspace is estimated using all available feature vectors (including both training and evaluation samples), reflecting a semi-supervised setting where unlabeled test data are availble during training. 
We then project the training data onto the top-$r$ components and fit the same logistic regression model. 
Unless otherwise specified, we fix $r=8$.

CPCR further applies a cross-fitted debiasing procedure with ridge regularization. 
The ridge parameter is coupled with the LR regularization through $\lambda = 1/C$, and we report the best performance over this shared grid.

\paragraph{Implementation details.}
All experiments are conducted with fixed random seeds to ensure reproducibility. 
We enforce deterministic behavior in both NumPy and PyTorch (including CUDA backends when available). 
Training is lightweight and completes within seconds on a single GPU (e.g., NVIDIA RTX 3060).

\paragraph{Model selection and reporting.}
For each method, we report the best test accuracy over the hyperparameter grid. 
Results are averaged over five independent random splits, and we report mean and standard deviation.








\end{document}